\PassOptionsToPackage{dvipsnames}{xcolor}
\documentclass{fairmeta}
\newif\ifarxiv\arxivtrue

 \usepackage{booktabs}
\usepackage[T1]{fontenc}
\usepackage{comment}
\usepackage[dvipsnames]{xcolor}
\usepackage{graphicx}
\usepackage{booktabs}
\usepackage{multirow}
\usepackage{amsmath}
\usepackage{amsfonts}
\usepackage{longtable}
\usepackage{enumitem}
\usepackage{xspace}
\usepackage{siunitx}
\usepackage[showseconds=false]{datetime2}
\usepackage{tikz}
\usepackage{pgfplots}
\pgfplotsset{compat=1.18}
\usepackage{standalone}
\usetikzlibrary{arrows.meta, decorations.pathreplacing, positioning, calc}
\usepackage{pifont}
\usepackage{placeins}
\usepackage{float}
\usepackage{hyperref}
\usepackage{subcaption}
\usepackage[font=small]{caption}

\usepackage{url}
\definecolor{darkgreen}{rgb}{0.0, 0.8, 0.13}

\def\relu{\texttt{RELU}\xspace}
\def\gelu{\texttt{GELU}\xspace}
\def\srelu{\texttt{squaredRELU}\xspace}
\def\prelu{\texttt{pRELU}\xspace}

\def\silu{\texttt{SILU}\xspace}
\def \Tchin {T\xspace}
\def \Wparam {LoPA Gating\xspace}
\def \Wfullrank {Standard Gating\xspace}
\def \Actcasting{Activation Casting\xspace}
\def \LoPcasting{LoPA Casting\xspace}
\def \Archcasting{Architecture Casting\xspace}

\def \RminSplus {\texttt{R-S+}\xspace}

\newcommand{\Base}{Base}
\newcommand{\LoPA}{LoPA}

\definecolor{dgreen}{RGB}{0,110,40}
\definecolor{dred}{RGB}{175,0,0}
\newcommand{\dpos}[1]{\textcolor{dgreen}{#1}}
\newcommand{\dneg}[1]{\textcolor{dred}{#1}}

\newif\ifshowrl\showrlfalse
\newcommand{\cmark}{\ding{51}}
\newcommand{\xmark}{\ding{55}}

\DeclareMathOperator{\sign}{sign}
\DeclareMathOperator*{\RMSNorm}{RMSNorm}

\makeatletter
\def\mypar{\@startsection{paragraph}{4}{\z@}{1.4em}{-1em}
  {\normalfont\normalsize\bfseries}}
\makeatother

\title{Model casting \& Low-parameter gating: \newline
towards more sparsely activated FFNs}

\author[*,1]{Maria Lomeli}
\author[*,1,2,3]{Antoine Groudiev}
\author[1]{Matthijs Douze}
\author[1]{Lo\"{\i}c Cabannes}
\author[1]{Pierre-Emmanuel Mazar\'e}
\author[1]{\newline  Fran\c{c}ois Fleuret}
\author[1]{Maximilian Beck}
\author[1]{Gergely Szilvasy} \author[1]{Naila Murray}
\author[1]{Herv\'e J\'egou}
\affiliation[1]{Meta FAIR}
\affiliation[2]{École Normale Supérieure -- PSL}
\affiliation[3]{École Normale Supérieure Paris-Saclay}

\contribution[*]{Equal contribution}
\correspondence{\email{marialomeli@meta.com}}
\date{\today}

\abstract{

This paper introduces \emph{model casting}, a mid-training recipe that drastically sparsifies the activations within the Feed-Forward Network (FFN) layer.
With this strategy, at inference time, we first compute the output of the gating matrix and, thanks to its high sparsity, we avoid computations with the two other matrices, reducing the FLOP count by up to $3\times$.
While this theoretical speedup is an upper bound, model casting translates into significant speedups both on CPU and GPU. \medskip \newline
We then introduce \Wparam, a new FFN design that increases the maximum theoretical speedup.
It is a low-FLOPs parameterization of the gating matrix that overcomes the $3\times$ cap by allocating fewer FLOPs and parameters to the gating matrix, compared to the two other FFN matrices that are sparsely activated.
\medskip
\newline
We consider two cases:
(i) we cast a pre-trained model with a sparsity inducing activation; (ii) we train with LoPA from scratch.
In all settings, we significantly outperform existing pruning solutions and regular \relu-fication.
For instance, at matched quality, we achieve a $3.2\times$ FLOP speedup  with \LoPcasting, against $1.6\times$ at best for competing methods top-$p$ and TEAL.
Using dedicated kernels, we achieve an actual $3.31\times$ speed-up on GPU at $90\%$ sparsity, past the $3\times$ ceiling of standard gating; \relu-fication, meanwhile, plateaus below $80\%$ sparsity.
}

\begin{document}
\maketitle

\addtocontents{toc}{\protect\setcounter{tocdepth}{-5}}

\ifarxiv
  \begin{figure}[h]
\centering
\includegraphics[width=0.9  \linewidth]{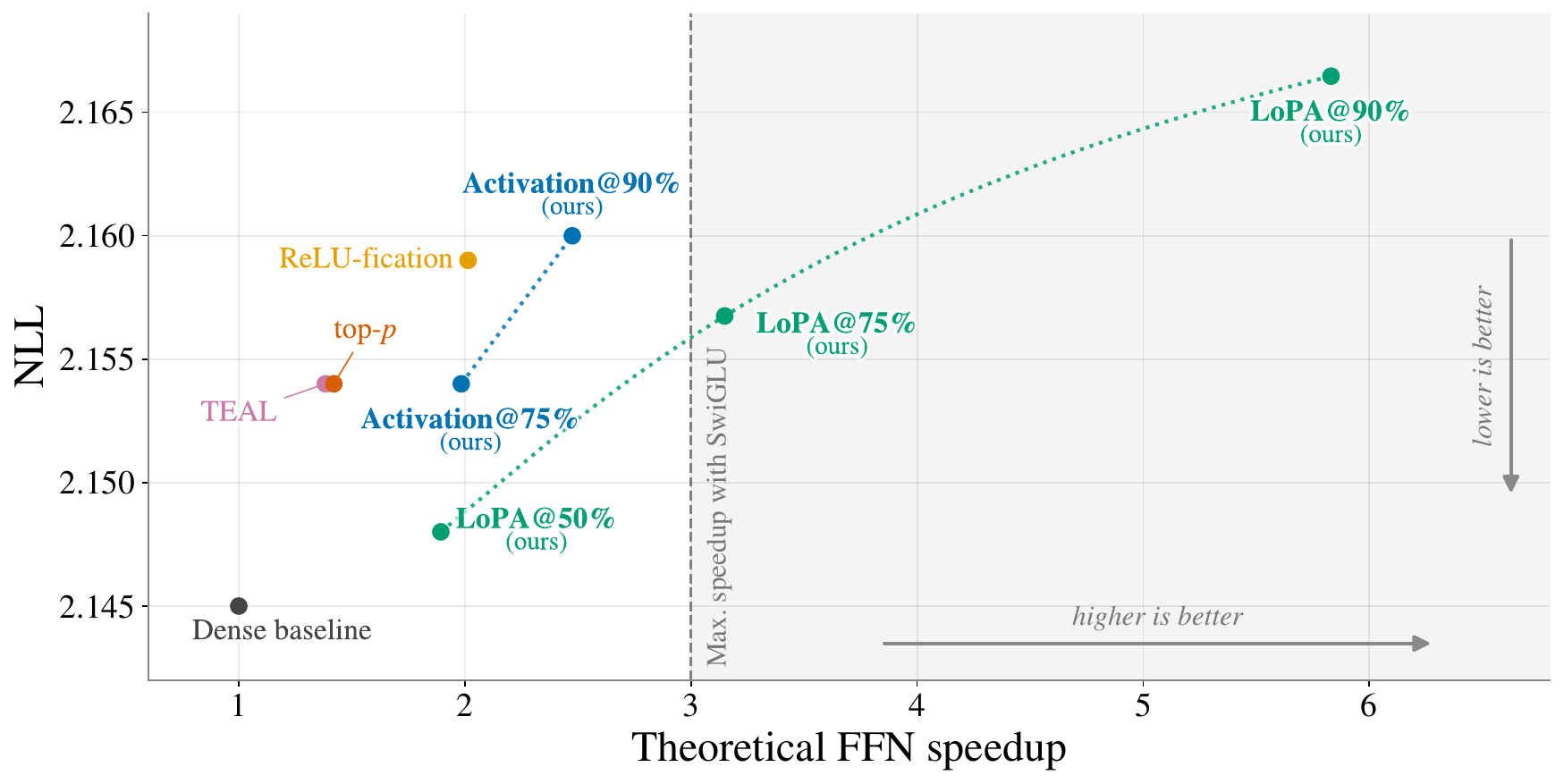}
\caption{
  \textbf{Model casting speedup at comparable quality,}
  \emph{vs} the baseline dense model (reference $\times1$ and NLL\,=\,$2.145$) and existing sparsification techniques: \relu-fication and the training-free top-$p$ and TEAL.
  At a $90\%$ sparsity target, \LoPcasting provides a $5.8\times$ speed-up for a $0.93\%$ NLL increase over the dense baseline.
  See detailed results in Tables~\ref{tab:model-casting-9T} and~\ref{tab:flop-posthoc_short}, and practical speed-up in Section~\ref{sub:speedups}.
  }
\label{fig:teaser}
\end{figure}

\fi

\section{Introduction}
\label{sec:introduction}

In current Large Language Model (LLM) architectures~\citep{Vaswani2017AttentionIA}, there exist two distinct memorization mechanisms. One is the short-term memory, determined by the  sequence at hand and instantiated as a sequence of keys and values produced in the attention mechanism.
The other one is the fixed parametric memory, which is populated at training time.
This memory is primarily contained in the part of the model that has the most parameters: the feed-forward (FFN) layers.

Both these memories involve some significant computational complexity. In particular, reducing the quadratic complexity of the attention mechanism is a major line of research in LLM optimization.
Although the FFN's inference cost is independent of sequence length, it dominates the complexity in most settings, unless the input context is extremely long.

In this paper, we focus on optimizing FFN inference speed.
We show how to fine-tune a pre-trained LLM so that its FFN post-activations become sparse enough to yield actual inference speedups,
against the popular belief that unstructured activation sparsity is unexploitable.
Our approach, which combines two main ingredients, trades large speedups for a much lower decrease in quality than existing methods, see Figure~\ref{fig:teaser}. Specifically, our contributions are as follows: \\[-1.5em]
  \begin{itemize}
  \item A mid-training strategy combining an activation swap with a sparsity-inducing loss, which effectively sparsifies FFN blocks without degrading model quality. This method is straightforward to implement: we replace the \silu activation with \RminSplus, a \relu-derived variant, and add an adaptive $\ell_1$ penalty. Unlike prior methods, the sparsity level is a target we set: we reach $90\%$ sparsity at a small NLL cost. \\[-1.1em]
  \item A new FFN design based on an efficient parameterization of $W_1$ that is combined with the activation swap strategy. It lowers the FFN FLOPs floor, enabling further efficiency gains. \\[-1.1em]
  \item Custom CPU and GPU kernels showing that the induced sparsity is practically exploitable: on a 5.8B-parameter model at $90\%$ sparsity, we obtain $1.53\times$ end-to-end CPU and $2.06\times$ FFN GPU speedups with standard gating, increasing to $1.88\times$ and $3.31\times$ when casting is combined with \Wparam, and to $3.90\times$ on GPU at $95\%$ sparsity. \\[-1.1em]
 \item Experiments showing that our approach is not tied to specific pre-trained models: model casting translates to Mixture-of-Experts, casting either the expert FFN blocks or the router that selects active experts. It also transfers to an off-the-shelf base model Qwen3-1.7B. \\[-1.1em]
\end{itemize}

\section{Related Work}
\label{sec:related_work}

\mypar{Emergent and trained-in activation sparsity.}
When a sparse activation such as \relu is used, trained transformers produce sparse intermediate activations without this being an explicit design choice~\citep{li2023the}. \citet{luo2025sparsing} show that the effect follows a power law in the amount of training data and that it is largely insensitive to model scale.
The switch to  \silu-gated designs~\citep{Dauetal17,Shazeer2020GLUVI} traded that property away: the \silu-gated FFNs that are now standard~\citep{jiang2023mistral7b,Riviere2024Gemma2I,Touvron2023LLaMAOA} have better performance but produce no exact zeros to exploit, while pre-training from scratch with \relu instead risks permanently dead neurons.
Recovering the sparsity afterwards is often treated during a mid-training stage: \citet{mirzadeh2024relu} swap the activation to \relu when fine-tuning \gelu or \silu models~\citep{zhang2022opt,touvron2023llama2}, and report much lower sparsity for gated FFNs than for non-gated ones.~\citet{lomeli2025stochasticactivations} prepare the network during pre-training by stochastically switching between \relu and \silu for negative values.
In both cases, the attainable sparsity is determined by the activation swap rather than directly controlled. In contrast, with model casting, we explicitly set the desired sparsity.

\mypar{Post-hoc sparsification.}
A complementary family exploits sparsity at inference time without retraining the base model: predictor-based \emph{contextual sparsity}, which selects an input-dependent subset of heads and neurons and underlies serving systems such as PowerInfer~\citep{liu2023deja,song2023powerinfer}; magnitude thresholding across all attention and MLP matrices, as in TEAL~\citep{Liuetal2025} and top-$p$~\citep{szatkowski2026universal}; calibrated per-layer thresholding of the \silu gate itself~\citep{lee2024cats}; and transform-then-threshold variants that first make the activations more compressible~\citep{Zhang2025RSparseRA,liu2025la}.
Such methods are bounded by the sparsity a pre-trained model happens to admit. For instance, \citet{szatkowski2026universal} report a \emph{critical sparsity}, the level reachable at $1\%$ quality degradation, of at most $40\%$ for an 8B model. None of the existing methods can prune the always-on gate, which is the cost that our \Wparam proposal removes.
The predictor-based schemes additionally target \relu models, where the MLP sparsity is already present, rather than the non-sparse gated FFNs we consider.
We compare against TEAL, top-$p$ and \relu-fication in Section~\ref{subsec:comparison-baselines}; their implementation details are provided in Appendix~\ref{sec:baselines}.

\mypar{Sparsity and efficiency by architecture.}
A third line of work changes the parameterization rather than the activation.
Since LoRA~\citep{hu2021loralowrankadaptationlarge}, low-rank factorizations have been used well beyond adaptation: to pre-train the FFN block with structured matrices~\citep{wei2024buildingefficientfoundationseffectively} or on the low-rank manifold of the product~\citep{mo2025parameter}, as a sum of low-rank and sparse matrices that is not restricted to a low-rank structure~\citep{han2024sltrain}, and as the diagonal-plus-low-rank structure of~\citet{chen2026structural}, which restores the universal approximation property.
\Wparam (Section~\ref{subsec:lopa-gating}) belongs to this family, but is used to make the \emph{gate} cheap rather than the whole block small.
Mixture-of-Experts models~\citep{Yang2024Qwen25TR,wei2024skyworkmoedeepdivetraining,deepseekv2,Jiang2024MixtralOE} instead place the sparsity in the routing, activating a subset of experts per token, conventionally through a softmax followed by top-$k$ selection~\citep{deepseekv2}; \citet{wang2025remoe} make that selection differentiable with \relu and an $\ell_1$ loss.
The two views meet: \citet{lee2024cats} observe that the \silu gate of a dense FFN can be read as a MoE-style router over hidden units, which is why the same casting mechanism applies to the FFN gate and to the MoE router alike, see Appendix~\ref{sec:moe-case-study} for further details.

\section{Method}
\label{sec:method}

\emph{Model casting} refers to changing part of a pre-trained model's architecture midway through training and continuing to train so that the model adapts to the change.
We consider three distinct setups: \\[-1.3em]
\begin{itemize}
    \item {\bf \Actcasting}: the model is pre-trained with a full matrix $W_1$ (\Wfullrank); at casting time, we replace the non-sparse activation by the new $\RminSplus$ activation (Section \ref{subsec:sparsity-exploitable-flops}), and mid-train the new model to magnify sparsity (Section~\ref{sec:model_casting}).  \\[-1.1em]
    \item {\bf \LoPcasting}: the model is pre-trained with the new \Wparam parameterization,
    see Section~\ref{subsec:lopa-gating}.
    Then we apply activation casting as in the first setup.
    The resulting model has the same number of parameters but is faster at inference. However, this setting requires the model to be trained from scratch with the \Wparam architecture. \\[-1.1em]
    \item {\bf \Archcasting}: we start from a pre-trained off-the-shelf model with \Wfullrank, see
    Section~\ref{sec:architecture_casting}. At casting time, the full $W_1$ matrix is converted to \Wparam and \Actcasting is applied at the same time. \\[-1.3em] 
\end{itemize}

\subsection{Preliminaries}
\label{subsec:sparsity-exploitable-flops}

\mypar{Gated FFN.} Transformers~\citep{Vaswani2017AttentionIA} have a number of blocks consisting of attention and feedforward components. We focus on FFN components with gated design, where,
on an input $x\in\mathbb{R}^D$, a gated FFN of hidden dimension $H>D$ computes as follows:
\begin{equation}
    y = W_2 \times (A(W_1 \times x) \odot (W_3 \times x))
    \textrm{  with }
    W_1, W_3 \in \mathbb{R}^{H\times D}
    \textrm{  and }
    W_2 \in \mathbb{R}^{D\times H},
\label{eq:ffn}
\end{equation}
where we assume column vectors. The symbol $\odot$ denotes element-wise multiplication.
The activation $A$ is either non-sparse, such as \silu or
\gelu, or sparse, such as \relu.
We consider single input vectors $x$, i.e. we do not batch inputs.

\begin{minipage}{0.61\linewidth}
\mypar{\RminSplus activation.} We consider the activation \RminSplus defined in the adjacent table, where $\sigma(x)\,{=}\,1/(1+\exp(-x))$ is the sigmoid function. It is a hybrid of \relu and \silu, equal to \silu on the positive side, thereby reducing the discrepancy introduced when casting the activation from \silu, while providing the sparsification behavior of \relu for $x<0$.
\end{minipage}
\hfill
\begin{minipage}{0.32\linewidth}
{\small
\begin{tabular}{p{1cm}c@{\quad\quad}c}
\toprule
       & $x<0$               & $x\geq 0$             \\
\midrule
\relu  & $0$                 & $x$                   \\
\silu  & $x \cdot \sigma(x)$ & $x \cdot \sigma(x)$   \\
\RminSplus  & $0$                 & $x \cdot \sigma(x)$   \\
\bottomrule
\end{tabular}
}
\end{minipage}

\mypar{FFN FLOP-ratio.} Each of the three matrices costs $HD$ multiply-adds per token, so a dense FFN costs $3HD$.
A zero entry is exploitable only if it is known before the matmul it is fed to: a zero in the gate $A(W_1x)$ is known after the $W_1x$ computation and before $W_3$, so it lets us skip one row of $W_3$ and one column of $W_2$, while $W_1$ is always computed in full.
Writing $\text{cost}(W)$ for the multiply-adds actually executed, we measure the reduction factor of FLOPs as
\begin{equation}
\text{FLOP-ratio}=\frac{3HD}{\text{cost}(W_1)+\text{cost}(W_2)+\text{cost}(W_3)}
\label{eq:flops_ratio}
\end{equation}
with standard gating and a sparsity fraction $s\in[0, 1]$, $\text{cost}(W_1)=HD$ and $\text{cost}(W_2)=\text{cost}(W_3)=(1-s)HD$, giving $\text{FLOP-ratio}=3/(3-2s)\in[1,3]$.

\subsection{Activation casting}
\label{sec:model_casting}

\mypar{Activation casting} is our simplest proposal:
We start from a checkpoint taken at a stage where the learning rate is stable, and replace the \silu activation of every FFN by \RminSplus. We then resume the training with a short warm-up and a stable phase so that the model adapts to the new activation, then decay the learning rate as usual. During this mid-training phase, we add the $\ell_1$ loss term to magnify sparsity.
Since \RminSplus coincides with \silu for $x\geq 0$, model casting only perturbs the negative half of the activation.
The alternative \relu-fication~\citep{mirzadeh2024relu}, by contrast, demands an adaptation on the whole data range, which has a larger impact on performance.

\mypar{Increasing sparsity with a regularizer.}
We measure the FFN sparsity (after mid-training with model casting) as the proportion of zeros in the post-activation gate output.
\citet{lomeli2025stochasticactivations} report more than 90\% sparsity with their \emph{stochastic activations} trained from scratch. 
However, they observed much lower sparsities when this stochastic activation is applied mid-training.

Hence, we further increase the sparsity by adding an $\ell_1$ sparsity loss for the hidden activations:
$$ \mathcal{L} = \mathcal{L}_{\text{CE}} +\frac{\lambda}{H}\sum_{\ell=1}^{L} \left| \mathbf{h}^{(\ell)} \right|_1$$
where $\mathcal{L}_{\text{CE}}$ is the standard per-token cross-entropy loss, $\mathbf{h}^{(\ell)} \in \mathbb{R}^H$ is the FFN activation at layer $\ell$, i.e. $A(W_1 x)$, and $|\cdot|_1$ is the $\ell_1$ norm (sum of absolute values).
Thus, there is a trade-off:
when $\lambda$ increases, the activation is sparser, but $\mathcal{L}_{\text{CE}}$  increases, degrading the model's generation quality.

\mypar{Adaptive $\ell_1$ loss.}
The relationship between the weight $\lambda$ and the sparsity $s$ depends on the model scale; see Appendix~\ref{app:l1_sweep}.
To set $s$, we adopt an \emph{adaptive} $\ell_1$ regularization \citep{wang2025remoe}:
given a target sparsity $s^\star$, $\lambda$ is initialized at $\lambda_0$ and at training step $i$ it is updated as
    $\lambda_{i+1} = \lambda_i \cdot \alpha^{\sign(s^\star-s_i)}$,
where $s_i$ denotes the average sparsity of all activations at step $i$ and $\alpha>1$ is a hyperparameter.
Hence, $\lambda$ increases if the sparsity is below the target and vice versa.
We set $\lambda_0=10^{-3}$ and \mbox{$\alpha=1.03$}.

\subsection{\Wparam}
\label{subsec:lopa-gating}

\begin{figure}[t]
    \centering
    \includegraphics[width=1.1\linewidth,page=4]{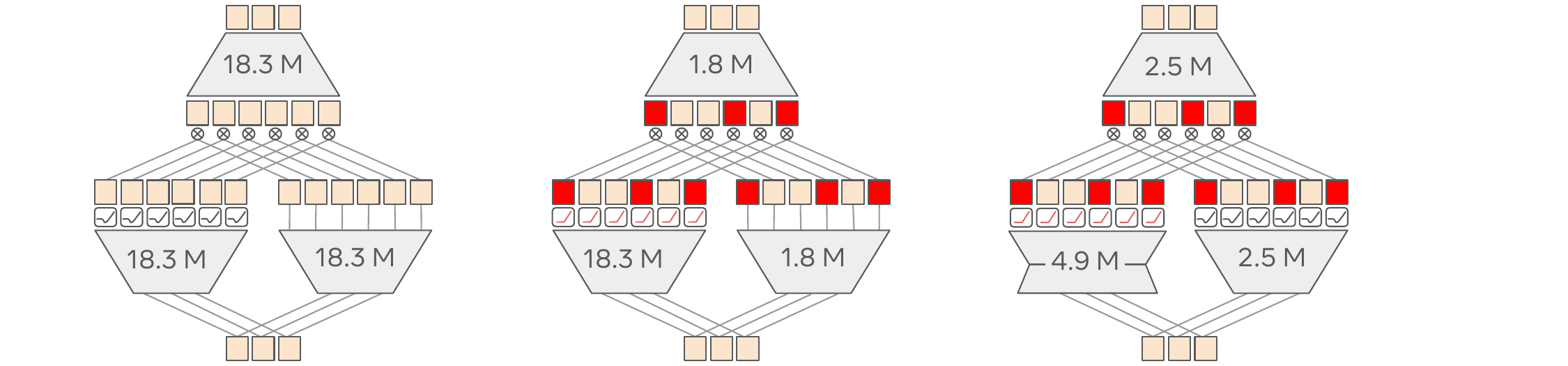}
    \par\vspace{-0.2em}
    \noindent\makebox[\linewidth]{
      \makebox[0.3\linewidth][c]{\hspace{-2.5em}\small\textbf{(a)} Dense SwiGLU}
      \makebox[0.3\linewidth][c]{\hspace{-3.5em}\small\textbf{(b)} Sparse SwiGLU}
      \makebox[0.3\linewidth][c]{\hspace{-3.5em}\small\textbf{(c)} Sparse \Wparam}
    }
    \caption{\textbf{Fewer FLOPs with sparsity and \Wparam.}
    FLOPs per linear layer at $D\,{=}\,2560$, $H\,{=}\,7168$ and rank $r\,{=}\,400$, with $90\%$ sparse $\RminSplus$ gates.
    In (c), the hidden dimension grows from $H$ to $H'{=}\,9787$ to keep a constant parameter count.
    The FLOPs reductions are due to sparsity in (b,c)  and to \Wparam in (c).
    }
    \label{fig:ffn_cost}
\end{figure}

The maximum theoretical FLOP-ratio is $3$ (Equation~\ref{eq:flops_ratio} with $s=1$), as we need to compute $W_1\times x$ to know where the zeros are.
We propose an architecture change to the FFN blocks to reduce this cost:
we make $\textrm{cost}(W_1)$ cheaper by decomposing it and reducing its number of parameters.
To keep the model's capacity fixed, we redistribute the saved parameters to $W_3$ and $W_2$, see Figure~\ref{fig:ffn_cost}(c).

\begin{figure}[t]
  \definecolor{nzcolor}{RGB}{52,118,196}      
  \definecolor{traincolor}{RGB}{214,90,52}    
  \definecolor{fixedcolor}{RGB}{150,150,150}  
  \tikzset{
  box/.style={draw, rounded corners, minimum height=8mm, minimum width=11mm,
              align=center, font=\small},
  op/.style={draw, circle, inner sep=0.5pt, minimum size=6mm, font=\small},
  flow/.style={-{Latex[length=2mm]}, thick},
  gridln/.style={gray!45, very thin},
  mat/.style={draw, thick, inner sep=1mm, align=center},
}
  \begin{minipage}[c]{0.49\textwidth}
\scalebox{1.13}{
    \begin{tikzpicture}[every node/.style={font=\small, transform shape=false}]
      \node[mat, fill=nzcolor!20, minimum width=13mm, minimum height=24mm,
            label=below:{\scriptsize $H\times D$}] (W1) {$W_1$};

      \node[right=1mm of W1] (eq) {$=$};

      \node[mat, fill=white, minimum width=13mm, minimum height=24mm,
            label=below:{\scriptsize $H\times D$},
            right=1mm of eq,
            path picture={
              \coordinate (gNW) at (path picture bounding box.north west);
              \coordinate (gNE) at (path picture bounding box.north east);
              \coordinate (gSW) at (path picture bounding box.south west);
              \coordinate (gSE) at (path picture bounding box.south east);
              \coordinate (gT1) at ($(gNW)!0.1667!(gNE)$);
              \coordinate (gT2) at ($(gNW)!0.3333!(gNE)$);
              \coordinate (gT3) at ($(gNW)!0.5!(gNE)$);
              \coordinate (gT4) at ($(gNW)!0.6667!(gNE)$);
              \coordinate (gT5) at ($(gNW)!0.8333!(gNE)$);
              \coordinate (gB1) at ($(gSW)!0.1667!(gSE)$);
              \coordinate (gB2) at ($(gSW)!0.3333!(gSE)$);
              \coordinate (gB3) at ($(gSW)!0.5!(gSE)$);
              \coordinate (gB4) at ($(gSW)!0.6667!(gSE)$);
              \coordinate (gB5) at ($(gSW)!0.8333!(gSE)$);
              \draw[gridln] (gT1) -- (gB1);
              \draw[gridln] (gT2) -- (gB2);
              \draw[gridln] (gT3) -- (gB3);
              \draw[gridln] (gT4) -- (gB4);
              \draw[gridln] (gT5) -- (gB5);
              \foreach \f in {0.0833,0.1667,0.25,0.3333,0.4167,0.5,0.5833,0.6667,0.75,0.8333,0.9167} {
                \draw[gridln] ($(gNW)!\f!(gSW)$) -- ($(gNE)!\f!(gSE)$);
              }
              \coordinate (gD1) at ($(gT1)!0.1667!(gB1)$);
              \coordinate (gD2) at ($(gT2)!0.3333!(gB2)$);
              \coordinate (gD3) at ($(gT3)!0.5!(gB3)$);
              \coordinate (gD4) at ($(gT4)!0.6667!(gB4)$);
              \coordinate (gD5) at ($(gT5)!0.8333!(gB5)$);
              \fill[fixedcolor!50] (gNW) rectangle (gD1);
              \fill[fixedcolor!50] (gD1) rectangle (gD2);
              \fill[fixedcolor!50] (gD2) rectangle (gD3);
              \fill[fixedcolor!50] (gD3) rectangle (gD4);
              \fill[fixedcolor!50] (gD4) rectangle (gD5);
              \fill[fixedcolor!50] (gD5) rectangle (gSE);
            }] (G) {\phantom{$G$}};
      \node[fill=none, inner sep=1mm] at (G.center) {$G$};

      \node[right=1mm of G] (plus) {$+$};

      \node[mat, fill=traincolor!35, minimum width=7mm, minimum height=24mm,
            label=below:{\scriptsize $H\times r$},
            right=1mm of plus] (U) {$U$};

      \node[right=1mm of U] (cdot) {$\cdot$};

      \node[mat, fill=traincolor!35, minimum width=18mm, minimum height=7mm,
            label=below:{\scriptsize $r\times D$},
            right=1mm of cdot] (V) {$V$};
    \end{tikzpicture}}
  
  \vspace{0.4em}
    \caption{\textbf{\Wparam\ parameterization}. The gating matrix is decomposed as $W_1\,{=}\,G+UV$, $U$ and $V$ are low-rank trainable matrices and $G$ is a non-trainable sparse matrix.
    The resulting $W_1$ is full rank, yielding much better results than $W_1=UV$ only, while having the same number of FLOPs and parameters. }
    \label{fig:loradiag}
  \end{minipage}
\hfill
\begin{minipage}{0.45\textwidth}
    \includegraphics[width=\linewidth]{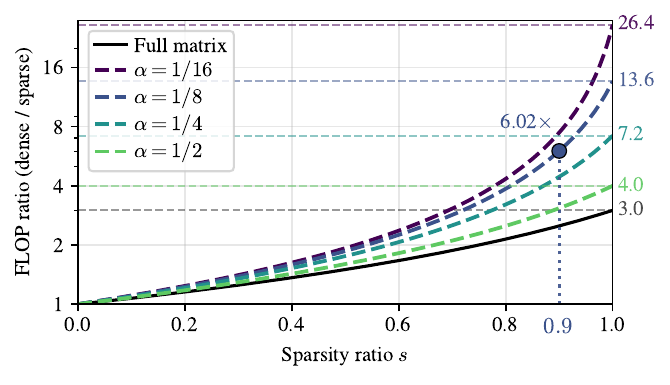}

\vspace{-0.5em}
    \caption{\textbf{FLOP-ratio for \Wparam}, as a function of the rank ratio $\alpha=r/D$ and of the sparsity ratio $s$. 
    At $\alpha=1/8$ and $90\%$ sparsity, the FFN FLOPs are reduced by $6.02\times$.
    }
    \label{fig:lora-theoretical-ratios}
    \end{minipage}
\end{figure}

\mypar{\Wparam parameterization.}
We represent $W_1$ as:
\begin{equation}
W_1 = G+UV
\textrm{  where }
U \in \mathbb{R}^{H\times r}
\textrm{  and }
V \in \mathbb{R}^{r\times D},
\label{eq:lopaapprox}
\end{equation}
where $G\in \mathbb{R}^{H\times D}$ is a non-trainable matrix with a single $1$ on each row, and $0$ elsewhere, see Figure~\ref{fig:loradiag}.
We call the full-rank $W_1$ version \emph{\Wfullrank} to distinguish it from \Wparam.

Adding the matrix $G$ prevents $W_1$ from becoming rank-deficient, similar to the diagonal-plus-low-rank parameterization of~\citet{chen2026structural}, for which a full-rank additive term restores the universal approximation property of the FFN.
Our $G$ generalizes that diagonal to any matrix with a single non-zero per row.
The design is justified empirically in  Appendix~\ref{sec:appendix-abl-wparam-design}:
\Wparam outperforms a low-rank gate ($W_1=UV$), and any fixed matrix with a single $\pm1$ per row performs equally well.
\Archcasting, described in the following, exploits this degree of freedom.

\mypar{\Wparam in practice.}

The rank $r$ controls the distribution of compute between the three matrices.
Lowering the rank makes inference cheaper but can degrade the validation NLL.
Noting $r=\alpha D$, where $\alpha\in(0, 1)$, we set $\alpha=1/8$ unless specified.
Replacing $W_1$ by the product of two matrices increases the depth of the network, making training at large scale unstable.
To overcome this, we add a Root Mean Square Normalization ($\RMSNorm$,  \cite{Zhang2019RootMS}) between the two low-rank terms:
  $W_1(x) = Gx + U \cdot \RMSNorm(Vx)$.

\mypar{Adaptive hidden dimension.}

To keep the FFN's number of parameters constant with \Wparam, we adapt the hidden dimension $H$ to $H'$.
Therefore, we solve for $H'$ so that $r(H'+D)+2H'D$ is equal to the standard parameter count $3HD$: $H'=D(3H-r)/(r+2D)$.

\mypar{Double activation.}
At this stage, \Wparam usually underperforms relative to \Wfullrank (see Table~\ref{tab:w1-design-ablation}).
Intuitively, reducing the capacity before the activation weakens the FFN's ability to use  its only non-linearity.
The \emph{double activation} avoids this: an activation function is applied to the outputs of \emph{both} $W_3$ and $W_1$, see Figure~\ref{fig:ffn_cost} (we cast only $W_1$'s activation).
This results in a slight performance increase compared to the \Wfullrank, see Appendix~\ref{sec:appendix-abl-wparam-design} for detailed ablations.

 \mypar{FLOP-ratio for \Wparam.}
  Omitting the negligible cost of the fixed $G$, the gate $W_1\,{=}\,UV$ costs $r(D+H')$, while $W_3$ and $W_2$ cost $(1-s)DH'$ each:
  \begin{equation}
    \text{FLOP-ratio}(s) = \frac{r(D+H') + 2DH'}{r(D+H') + 2(1-s)\,DH'}.
  \end{equation}
  In our scaling ladder, $H/D$ is kept constant at $8/3$ across model sizes.
  When $s\to1$ the FLOP-ratio floor depends solely on $\alpha\,{=}\,r/D$, see Figure~\ref{fig:lora-theoretical-ratios}.
  For example, with $\alpha\,{=}\,1/8$ \Wparam's FLOP-ratio becomes 13.5, to be compared to just 3 for \Wfullrank.

\subsection{Architecture casting}
\label{sec:architecture_casting}

Model casting can be applied on top of a pre-trained model:
we refer to this case as \emph{\Archcasting}, in which the full gating matrix $W_1$ is replaced with \Wparam: at casting time, we replace the $W_1$ matrix with the low-rank expression from Equation~\ref{eq:lopaapprox}, where the matrices are chosen to approximate the current $W_1$.
We then resume the training of the model, which adapts to both the $\RminSplus$ activation and the new architecture.
For \Archcasting, we do not introduce an RMSNorm between $U$ and $V$, nor an activation after $W_3$, in order to keep the cast model as close as possible to the original checkpoint.
Note that replacing the full $W_1$ matrix by this parameterization reduces the number of FFN parameters (since we cannot adapt the hidden dimension $H$), which degrades the model's performance as expected.

\mypar{\Wparam approximation.}

To approximate matrix $W_1\in\mathbb{R}^{H\times D}$ with a \Wparam FFN at a fixed rank $r$, we solve:
\begin{equation}
    \min_{M\in \mathcal{C}, U\in\mathbb{R}^{H\times r}, V\in\mathbb{R}^{r\times D}} \lVert W - M - UV \rVert_F^2 
\end{equation}
where we choose $\mathcal{C}\subset \{0, -1, +1\}^{H\times D}$ to have exactly one nonzero entry per row (see Appendix~\ref{sec:appendix-abl-wparam-design} for alternative designs).
Solving this optimization is intractable, so we resort to an alternating minimization strategy, which
guarantees to decrease the objective: in turn we
\begin{itemize}
\item
    fix $U, V$: the optimal position and sign of the $\pm1$ in each row of $M$ are given by the largest-magnitude entry of the row in $W-UV$
\item
    fix $M$: the optimal $UV$ is the truncated Singular Value Decomposition (SVD) of $W-M$.
\end{itemize}
Since $\mathcal{C}$ is discrete, it converges in a finite number of iterations (a few steps in practice).
We start the optimization by setting $U=V=0$ and updating $M$ first.

\section{Experimental Setup}
\label{sec:experimental_setup}

\mypar{Model architecture}
\label{sec:dense_architecture}
We train dense decoder-only models.
The model architecture is derived from Llama3~\citep{dubey2024llama}.
We use the Llama3 tokenizer, a fast Byte-Pair Encoding tokenizer implemented with TikToken, with a  vocabulary of 128,000 regular tokens as well as 256 reserved tokens.
The transformer blocks use grouped-query attention~\citep{ainslie-etal-2023-gqa}, RMSNorm~\citep{Zhang2019RootMS} prenormalization, rotary positional encoding (RoPE)~\citep{Su2021RoFormerET} with $\theta\,{=}\,5 \cdot10^{5}$ and document causal masking.

\mypar{Training schedule}
\label{sec:wsd}

The Chinchilla scaling law~\citep{hoffmann2022training} states that the minimal training cost to obtain a given loss is reached when the number \Tchin of training tokens is roughly $20\times$ the number of model parameters (20 TPP).
We use this \Tchin as the unit for the training duration,
because it is independent of training hours (hardware dependent) or optimization steps (depends on batch size), and can be compared over model sizes.
To improve the inference-time accuracy, we consider training regimes longer than  1\Tchin,
following recent practice, e.g., \cite{dubey2024llama} or \cite{Riviere2024Gemma2I}.
We distinguish the pre-training (PT) from the mid-training (MT) budget, i.e. the number of tokens spent on model casting after the activation swap, excluding the final decay.

The PT phase lasts 4\Tchin.
Since the model needs to adapt to the activation change, we reset the optimizer state when starting MT.
The MT is divided into a three-phase WSD learning rate schedule:
\begin{itemize}
    \item {\bf W}arm-up: The learning rate linearly increases from 0 to $\gamma$ over $T/20$ tokens.
    This base learning rate $\gamma$ and the batch size are determined by scaling laws  (Appendix~\ref{sec:ladder}); 
    \item {\bf S}table: the learning rate is maintained at $\gamma$ until reaching \Tchin, 2\Tchin, 4\Tchin or 7\Tchin training tokens.
    The loss decreases slowly during this phase.
    \item {\bf D}ecay: the learning rate is decreased with a Cosine schedule over \Tchin tokens.
    The loss gain during this phase (the ``accuracy boost'') is roughly constant.
\end{itemize}
The WSD schedule allows us to quantify the extra amount of tokens required to reach the same training loss as the dense baseline.
Unless specified, we perform 4\Tchin pre-training, 4\Tchin W+S and 1\Tchin decay.

\mypar{Dense baseline}
The baseline is a vanilla \silu-based model that continues from the same pre-trained checkpoint, using the same MT budget and WSD schedule, but without an activation swap or $\ell_1$ loss. The comparison is therefore matched on both pre-training and mid-training compute.

\mypar{Metrics.}
We measure the sparsity and Negative Log-Likelihood (NLL), both computed on a held-out validation set.
The NLL on a validation set is less noisy than the training loss because it is evaluated on the same set of sequences for all models.
We average the sparsity $s$ over all FFN hidden activations.
This is the sparsity of the gate post-activation $A(W_1 x)$.

\mypar{Scaling ladder}

The scaling ladder starts with a set of eleven pre-trained models spanning nearly two orders of magnitude in size and complexity, ranging from 30 million to 5.79 billion non-embedding parameters~\citep{SziFayetal2026}.
We use the scaling ladder models for mid-training with model casting with $\ell_1$ and for \Wparam.
The AdamW optimizer~\citep{loshchilov2017decoupled} learning rate is tuned for the WSD schedule.
Each model in the ladder is trained for a total duration of 8\Tchin, see
Appendix~\ref{sec:ladder} for details.

\section{Results}
\label{sec:results}

\subsection{Model casting}

\mypar{Impact on NLL.}
Table~\ref{tab:model-casting-9T} shows model casting results during mid-training.
We compare the three setups (Activation, LoPA, and Architecture casting); for each setup, we train models both with and without the sparsifying $\ell_1$ loss, along with a reference run that uses the \silu activation throughout the entire training duration.
Each model is pre-trained for $4\Tchin$, cast for $4\Tchin$, and annealed for $1\Tchin$ with a decaying learning rate.
Activation and LoPA casting coupled with the $\ell_1$ loss achieve sparsity levels close to 90\%, while keeping the NLL degradation below 1\%.
Adding an $\ell_1$ loss slightly degrades the performance, but increases the sparsity level to the 90\% target, enabling FLOPs speedups up to $5.83\times$ in \LoPcasting.
As \Archcasting reduces the parameter count, it leads to a larger degradation of the NLL; however, Appendix~\ref{sec:arch-cast-vs-reduced} (Table~\ref{tab:arch-cast-vs-reduced}) shows that \Archcasting's NLL is on par with models with equivalent parameter counts.
We show that these findings are consistent at larger model scales in Appendix~\ref{sec:level13-nll}, where we observe smaller relative NLL degradations for 5.79B models.
We also evaluate the mid-trained models' downstream performance on a suite of 12 short-context tasks and compare them against a control given the same mid-training budget without model casting. \Actcasting costs $0.4$ points of the Core-12 average at a $90\%$ sparsity target, and $1.6$ points on Qwen3-1.7B; see Appendix~\ref{app:short_context}.

\begin{table}
    \centering
    {\small
    \caption{
    \textbf{Model casting enables significant FLOPs speedups at low NLL impact.}
    Results for 0.81B-parameter models (level 8 in Table~\ref{tab:ladder}). 
    $\Delta$NLL is computed relative to the first row in \textbf{bold}.
    We indicate the Base or LoPA architecture with the activation [in square brackets].
        \label{tab:model-casting-9T}
    }
    \begin{tabular}{lcccc}
        \toprule
        Setup (pre-training $\rightarrow$ mid-training) & NLL & $\Delta$NLL & Sparsity & Speedup \\
        \midrule
       \multicolumn{5}{l}{\textit{Baselines: mid-training and \relu-fication}}\\
        \qquad \Base[\silu{}] $\to$ \Base[\silu{}]~{\small(dense baseline)} & \textbf{2.145} & --- & --- & $1.00\times$ \\
         \qquad \Base[\silu{}] $\to$ \Base[\relu{}]~{\small(\relu-fication)} & 2.159 & $+0.65\%$ & 76\% & $2.00\times$ \\
        \midrule
        \multicolumn{5}{l}{\textit{(a) \Actcasting}}\\
        \qquad \Base[\silu{}] $\to$ \Base[\RminSplus{}]        & 2.153 & $+0.36\%$ & 69\% & $1.84\times$ \\
        \qquad \Base[\silu{}] $\to$ \Base[\RminSplus, $\ell_1$@90\%] & 2.160 & $+0.68\%$ & 89\% & $2.48\times$ \\
        \addlinespace
        \multicolumn{5}{l}{\textit{(b) \LoPcasting}}\\
        \qquad \LoPA[\silu{}] $\to$ \LoPA[\silu{}]          & 2.150 & $+0.24\%$ & --- & $1.00\times$ \\
        \qquad \LoPA[\silu{}] $\to$ \LoPA[\RminSplus{}]        & 2.148 & $+0.13\%$ & 51\% & $1.89\times$ \\
        \qquad \LoPA[\silu{}] $\to$ \LoPA[\relu{}]& 2.149 & $+0.19\%$ & 57\% & $2.11\times$ \\
        \qquad \LoPA[\silu{}] $\to$ \LoPA[\RminSplus, $\ell_1$@90\%] & 2.165 & $+0.93\%$ & 89\% & $5.83\times$ \\
        \addlinespace
        \multicolumn{5}{l}{\textit{(c) \Archcasting ($-21\%$ parameters and max FLOPs)}}\\
        \qquad \Base[\silu{}] $\to$ \LoPA[\silu{}]            & 2.180 & $+1.63\%$ & --- & $1.39\times$ \\
        \qquad \Base[\silu{}] $\to$ \LoPA[\RminSplus{}]       & 2.186 & $+1.90\%$ & 73\% & $4.21\times$ \\
        \qquad \Base[\silu{}] $\to$ \LoPA[\relu{}]          & 2.194 & $+2.27\%$ & 78\% & $4.87\times$ \\
        \qquad \Base[\silu{}] $\to$ \LoPA[\RminSplus, $\ell_1$@90\%]& 2.193 & $+2.20\%$ & 90\% & $7.94\times$ \\
        \bottomrule
    \end{tabular}}
\end{table}

\mypar{Catch-up duration.}
Another way of looking at the training process is to measure how many additional training tokens are needed to recover the model's pre-casting NLL.
We report this catch-up duration across model scales in Appendix~\ref{sec:catchup} (Table~\ref{tab:catchup}).
Recovering the pre-casting NLL takes at most $1.25\Tchin$ without the $\ell_1$ loss ($0.5$--$1.0\Tchin$ for \LoPcasting); driving the model to $90\%$ sparsity with the $\ell_1$ loss raises this to $1.5$--$2.75\Tchin$ (2.75\Tchin for the largest \LoPcasting model).
The catch-up duration is roughly constant across model scales.

\mypar{Mixture-of-Experts.}
Model casting also applies to MoE models, which is another way of performing sparse inference.
We evaluate on the expert FFN blocks of an eight-expert, top-$2$ model during mid-training for 4\Tchin, against a \silu{} top-2 model given the same mid-training budget (NLL $1.981$).
With a light $\ell_1$ penalty ($\lambda=10^{-3}$), the model reaches $59.0\%$ activation sparsity for $+0.033$ NLL, and targeting $90\%$ sparsity reaches $88.6\%$ for $+0.083$ NLL.
By comparison, switching to top-$1$ routing --- the standard architectural route to sparsity --- ends up at a \emph{worse} NLL ($2.073$ vs $2.064$) after the same mid-training budget, while yielding no activation sparsity at all.
Thus, MoE and \Actcasting can be combined.
Appendix~\ref{sec:moe-case-study} provides the full experimental setup, an alternative router that selects active experts through casting, and additional ablations; see Tables~\ref{tab:moe_sparsity_nll},~\ref{tab:nll_sparsity_cpt_moe} and~\ref{tab:moe_8e_top1_vs_top2}.

\mypar{Compatibility with quantization and post-training.}
In Appendix~\ref{sec:qat} we show that  quantization-aware training can be applied on top of models sparsified with model casting.
Combining both, we obtain a $1.57\%$ degradation of the NLL in \texttt{int8}. 
Our low-precision kernels, designed to exploit the sparsity, achieve an $8.13\times$ speedup at 90\% sparsity.
Model casting also survives post-training: LCFT and SFT cost model casting variants about as much as the dense control (Appendix~\ref{sec:post-training}).

\mypar{Additional ablations.}
We validate some of our design choices in the appendix.
An $\ell_1$ loss \emph{without} model casting cannot induce significant sparsity and degrades performance, and pre-training with \silu and $\ell_1$ before casting worsens the final sparsity (see Appendices~\ref{sec:only_l1} and~\ref{sec:abl-l1-pt}).
Among alternative casting activations (Appendix~\ref{sec:diff_activations}), \silu and \gelu give the best NLL, Shifted \relu~\citep{mirzadeh2024relu,qiu2017frelu} gives the highest sparsity at a high NLL cost. \RminSplus keeps the NLL close to \silu at moderate sparsity; the squared \relu advocated by \citet{Zhang2024ReLU2WD} when training from scratch gives the \emph{lowest} exact-zero sparsity in our model casting setting.

\subsection{Comparison to prior FFN sparsification methods}
\label{subsec:comparison-baselines}

We compare model casting with i) two training-free FFN sparsification approaches, top-$p$~\citep{szatkowski2026universal} and TEAL~\citep{Liuetal2025} and ii) another fine-tuning approach, \relu-fication~\citep{mirzadeh2024relu}.
See Appendix~\ref{sec:baselines} for the implementation details.
Every FLOP ratio we report is normalized to an equal-parameter dense standard-gating \silu model. For the training-free approaches, we further mid-train the pre-trained 4\Tchin model for the same total duration.

\begin{table}[t]
  \centering
  \small

    \caption{\textbf{Trained-in sparsity gives a larger FLOP speedup than
    post-hoc pruning at matched quality.} FFN theoretical FLOP speedup
    ($1/\text{FLOP-ratio}$; dense $=1\times$, higher is better), at batch
    size~1, with all methods matched at the reference NLL.
    See Appendix~\ref{sec:baselines} for other training budgets.}
   \begin{tabular}{l c c c c}
    \toprule
     & & \multicolumn{3}{c}{Theoretical FLOP speedup $\uparrow$} \\
    \cmidrule(lr){3-5}
    FFN Architecture & Ref. NLL\,$\downarrow$ & Model casting @75\% (ours) & top-$p$ & TEAL \\
    \midrule
    \Wfullrank & 2.154 & \textbf{2.0} & 1.4 & 1.4 \\
    \Wparam    & 2.153 & \textbf{3.2} & 1.3 & 1.3 \\
    \bottomrule
    \end{tabular}
  \label{tab:flop-posthoc_short}
\end{table}

\mypar{Training-free approaches.}
In Table~\ref{tab:flop-posthoc_short}, 
we match all methods with a common reference NLL (the validation NLL of model casting at 75\% sparsity) and compare the corresponding FLOP speedup.
Model casting results in higher sparsity levels and FLOPs speedups than post-hoc approaches:
$2$--$3.2\times$
against $1.3$--$1.6\times$ for top-$p$ and TEAL.
Moreover, the advantage grows for longer schedules, see
Table~\ref{tab:flop-posthoc} in Appendix~\ref{sec:baselines}.

\mypar{Trained approaches.}
Table~\ref{tab:model-casting-9T} compares our method versus \relu-fication. Model casting reaches higher sparsity at a larger FLOP speedup.
The full comparison across mid-training budgets is in Table~\ref{tab:flop-trained} of Appendix~\ref{sec:baselines}.
For \Actcasting, this comes with an NLL difference below $0.01$ nats; for \LoPcasting the $90\%$ target costs a small NLL increase ($0.01$--$0.04$) but the model is substantially cheaper.
Moreover sparsity is a target we set, so casting reaches around $90\%$ in every setting, whereas \relu-fication offers no comparable control mechanism.

\mypar{Casting an off-the-shelf checkpoint.}
Table~\ref{tab:qwen3-17b-model-casting} applies \Actcasting and \Archcasting to Qwen3-1.7B-Base: both reach more than $89\%$ sparsity, and \Archcasting incurs $0.062$ higher NLL than \Actcasting but removes $294$M parameters.
Mid-training details, the learning-rate sweep and downstream evaluations are in Appendix~\ref{sec:qwen3}.

\begin{table}[t]
    \centering
    \caption{Qwen3-1.7B base model mid-trained with model casting, $\gamma\,{=}\,7.4{\times}10^{-5}$.
        The 0\Tchin row is the pretrained 
        checkpoint that we try to recover with sparse models. 
        ``Decay'' indicates if the 1\Tchin cosine annealing is applied.
    }\vspace{-0.5em}
{\footnotesize
      \begin{tabular}{lccccc}
        \toprule
        Method & Budget & Decay & Sparsity (\%) & NLL& Speedup \\
        \midrule
        Pretrained (no casting) & 0\Tchin & ---  & --- & \textbf{1.843} & $1.00\times$ \\
        \midrule
                             & 1\Tchin & \xmark  & 89.6 & 1.912 & $2.48\times$ \\
        \Actcasting          & 2\Tchin & \xmark  & 89.6 & 1.908 & $2.48\times$ \\
        PT $\rightarrow$ Base[\RminSplus, $\ell_1$@90\%]
                             & 4\Tchin & \xmark  & 89.6 & 1.907 & $2.48\times$ \\
                             & 5\Tchin & \cmark & 89.5 & 1.875 & $2.48\times$ \\
        \midrule
                             & 1\Tchin & \xmark  & 89.6 & 2.020 & $8.01\times$ \\
        \Archcasting         & 2\Tchin & \xmark  & 89.7 & 1.991 & $8.05\times$ \\
        PT $\rightarrow$ LoPA[\RminSplus, $\ell_1$@90\%]
                             & 4\Tchin & \xmark  & 89.7 & 1.971 & $8.05\times$ \\
                             & 5\Tchin & \cmark  & 89.6 & 1.937 & $8.01\times$ \\
        \bottomrule
      \end{tabular}
      }
      \label{tab:qwen3-17b-model-casting}
\end{table}

\subsection{CPU/GPU optimization for FFNs with sparse activations}
\label{sub:speedups}

Exploiting sparsity is mainly useful at LLM generation time, during single-token autoregressive decoding.
With batched computation, the sparsity patterns do not coincide between batch elements. This reduces the sparsity as the rows/columns of $W_3$ and $W_2$ must be loaded for the union of the inverse sparsity masks.

\mypar{CPU and GPU kernels.}
On CPU, we use a sparsity-aware AVX-512 kernel.
On GPU, we implement a family of Triton kernels that fuse the $W_2$ and $W_3$ projections and skip the inactive neurons.
The exact blocking and tiling scheme is selected according to the sparsity.
All results are for float32 computations, see GPU results for bfloat16 in Appendix~\ref{sec:timing-bf16}.
On the GPU, overheads such as KV-cache management, Python interpretation, attention masks, and RoPE account for a larger fraction of runtime.
These overheads are less significant on larger models or carefully optimized software stacks.
We plot rooflines corresponding to incompressible costs in order to assess the tightness of our implementations.
On CPU, we plot the other operations, which account for the time spent outside of the FFN and that is  independent of sparsity.
On GPU, we plot the memory roofline, i.e. the speedup that would be attained by just loading the weights from High Bandwidth Memory.

\begin{figure}[t]
    \centering
    \includegraphics[width=\linewidth]{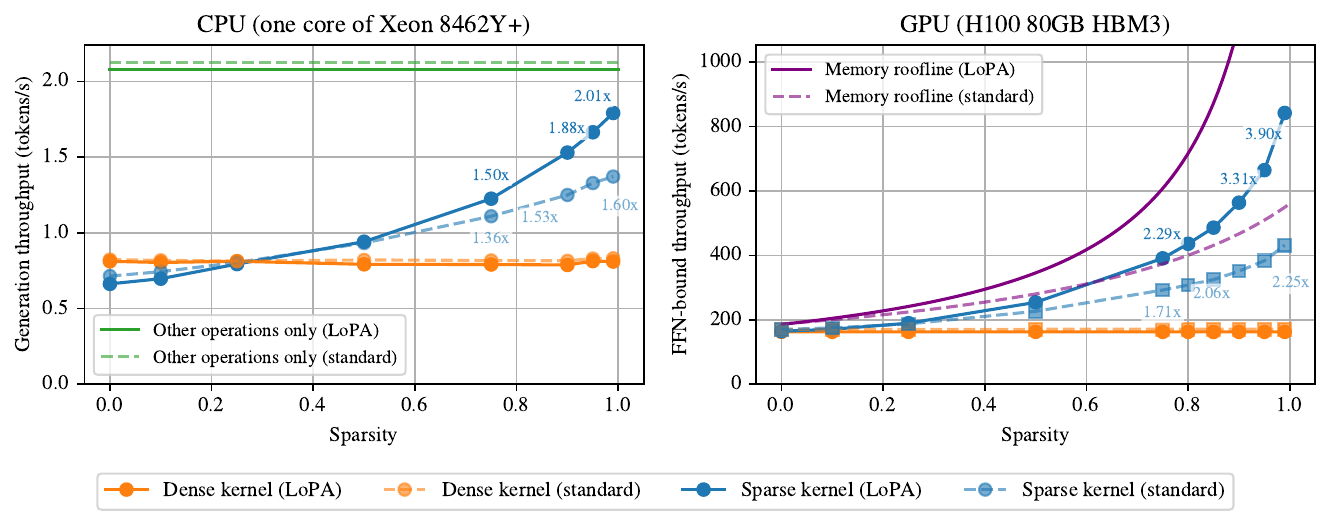}
    \caption{
    \textbf{Sparsity translates to wall time improvements on CPU and GPU.}
    Throughput as a function of activation sparsity, for a 5.79B non-embedding parameters model (level 13 in the scaling ladder from Table~\ref{tab:ladder}) in float32.
CPU times are end-to-end; GPU times cover the FFN only, with the memory-bandwidth roofline shown as an upper bound.
At 90\% sparsity, the speedup on GPU is $2.06\times$ for \Wfullrank and $3.31\times$ for \Wparam.
}
    \label{fig:runtimes}
\end{figure}

\mypar{Wall-clock speedups.}
Figure~\ref{fig:runtimes} reports wall-clock timings: end-to-end on the CPU, and FFN-only on the GPU, in contrast to the theoretical FLOP speedups reported in Section~\ref{subsec:comparison-baselines}.
On the CPU, 90\% sparsity directly translates into a practical $1.53\times$ wall-clock speedup for \Wfullrank, and $1.88\times$ for \Wparam.
For the GPU, the wall-clock speedup of the FFN inference is around $2.06\times$ at 90\% sparsity for \Wfullrank, and $3.31\times$ for \Wparam.
At 95\% sparsity on the GPU, \Wparam achieves $3.90\times$ (while \Actcasting is limited to 3$\times$).

\section{Conclusion}
\label{sec:conclusion}

\citet{szatkowski2026universal} define the ``critical sparsity'' to be the sparsity attainable at a $1\%$ quality drop, and estimate it as below $40\%$ for an 8B LLM.
We show that for the model casting introduced in this work, this number is closer to 90\%, and can be set as a target at training time.
Building on this, we introduced \Wparam, a novel FFN block that allocates the compute between the three FFN matrices to substantially improve the frontier of sparsity acceleration.
We proposed \Archcasting, to convert a pre-trained model (for example Qwen) to \Wparam,
and model casting applies to the experts of MoE models as well.
Our models achieve up to $3.3\times$ wall-clock FFN speedups compared to standard dense designs, while keeping the NLL degradation below $1\%$.

\FloatBarrier
\newpage

\appendix
\raggedbottom

\addtocontents{toc}{\protect\setcounter{tocdepth}{2}}  
\renewcommand{\contentsname}{Appendix Contents}
\begingroup
\makeatletter
\let\appendixoriginallsection\l@section
\renewcommand*{\l@section}[2]{
  \begingroup
  \let\addvspace\@gobble
  \appendixoriginallsection{#1}{#2}
  \endgroup
}
\makeatother
\tableofcontents
\endgroup

\section{Ablation on the \Wparam design}
\label{sec:appendix-abl-wparam-design}

\begin{figure}[ht]

    \definecolor{nzcolor}{RGB}{52,118,196}      
    \definecolor{traincolor}{RGB}{214,90,52}    
    \definecolor{fixedcolor}{RGB}{150,150,150}  
    \definecolor{grpA}{RGB}{120,170,220}
    \definecolor{grpB}{RGB}{240,180,90}
    \definecolor{ldA}{RGB}{214,40,40}
    \definecolor{ldB}{RGB}{235,140,30}
    \definecolor{ldC}{RGB}{180,160,20}
    \definecolor{ldD}{RGB}{60,160,80}
    \definecolor{ldE}{RGB}{40,140,190}
    \definecolor{ldF}{RGB}{150,80,185}
    \tikzset{
      box/.style={draw, rounded corners, minimum height=8mm, minimum width=11mm,
                  align=center, font=\small},
      op/.style={draw, circle, inner sep=0.5pt, minimum size=6mm, font=\small},
      flow/.style={-{Latex[length=2mm]}, thick},
      gridln/.style={gray!45, very thin},
    }
    \newcommand{\cellfill}[3]{\fill[#3] (#1,#2) rectangle ++(1,1);}

  \centering
  \begin{subfigure}[b]{0.32\linewidth}
    \centering
    \begin{tikzpicture}[scale=0.3]
      \cellfill{0}{11}{fixedcolor!60}   
      \cellfill{1}{10}{fixedcolor!60}   
      \cellfill{2}{9}{fixedcolor!60}    
      \cellfill{3}{8}{fixedcolor!60}    
      \cellfill{4}{7}{fixedcolor!60}    
      \cellfill{5}{6}{fixedcolor!60}    
      \draw[gridln] (0,0) grid (6,12);
      \draw[thick] (0,0) rectangle (6,12);
    \end{tikzpicture}
    \caption{Low-rank + Identity: $I_{\text{pad}}$}
    \label{fig:fixed-id}
  \end{subfigure}
  \hfill
  \begin{subfigure}[b]{0.32\linewidth}
    \centering
    \begin{tikzpicture}[scale=0.3]
      \cellfill{0}{11}{fixedcolor!60} \cellfill{0}{10}{fixedcolor!60}
      \cellfill{1}{9}{fixedcolor!60}  \cellfill{1}{8}{fixedcolor!60}
      \cellfill{2}{7}{fixedcolor!60}  \cellfill{2}{6}{fixedcolor!60}
      \cellfill{3}{5}{fixedcolor!60}  \cellfill{3}{4}{fixedcolor!60}
      \cellfill{4}{3}{fixedcolor!60}  \cellfill{4}{2}{fixedcolor!60}
      \cellfill{5}{1}{fixedcolor!60}  \cellfill{5}{0}{fixedcolor!60}
      \draw[gridln] (0,0) grid (6,12);
      \draw[thick] (0,0) rectangle (6,12);
    \end{tikzpicture}
    \caption{Low-rank + Diagonal: $G$}
    \label{fig:fixed-diag}
  \end{subfigure}
  \hfill
  \begin{subfigure}[b]{0.32\linewidth}
    \centering
    \begin{tikzpicture}[scale=0.3]
      \cellfill{0}{11}{fixedcolor!60} \cellfill{0}{5}{fixedcolor!60}
      \cellfill{1}{10}{fixedcolor!60} \cellfill{1}{4}{fixedcolor!60}
      \cellfill{2}{9}{fixedcolor!60}  \cellfill{2}{3}{fixedcolor!60}
      \cellfill{3}{8}{fixedcolor!60}  \cellfill{3}{2}{fixedcolor!60}
      \cellfill{4}{7}{fixedcolor!60}  \cellfill{4}{1}{fixedcolor!60}
      \cellfill{5}{6}{fixedcolor!60}  \cellfill{5}{0}{fixedcolor!60}
      \draw[gridln] (0,0) grid (6,12);
      \draw[thick] (0,0) rectangle (6,12);
    \end{tikzpicture}
    \caption{Low-rank + Tiled Identity: $T$}
    \label{fig:fixed-tid}
  \end{subfigure}
  \caption{\textbf{Illustration of the non-learnable component.}
  We depict, for $H=12$ and $D=6$, the fixed component $M\in\mathbb{R}^{H\times D}$ of three low-rank variants.
  (a) Low-rank + Identity uses the padded identity $I_{\text{pad}}$.
  (b) Low-rank + Diagonal uses the repeated diagonal $G$.
  (c) Low-rank + Tiled Identity uses the tiled identity $T$, that is
  $(Tx)_i=x_{\,i\bmod D}$. All non-zeros are fixed $1$s (gray).
  The last two variants, or any permutation of those, correspond to what we call LoPA. The simple identity $I_{\text{pad}}$ differs from LoPA as some rows only contain zeros.
}
  \label{fig:fixed-matrices}
\end{figure}
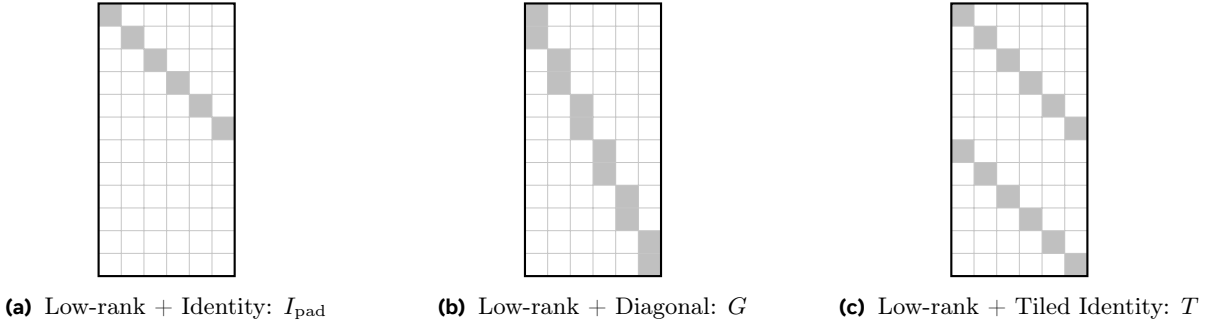

In order to demonstrate the effectiveness of the \Wparam architecture as an under-parameterized matrix design, we compare models with alternative architectural variations.
Unless stated otherwise, $W_1$ is parameterized using \Wparam: the LoPA variant consisting of low-rank and fixed diagonal-like components, with double activation (activations on the outputs of both $W_1$ and $W_3$), and the hidden dimension of the FFN blocks is increased in order to match the number of parameters of the baseline (as described in Subsection~\ref{subsec:lopa-gating}).
All models are trained with $\silu$ for a duration of $8\Tchin$, without any sparsity loss.
Their specific characteristics are as follows:
\begin{itemize}
    \item {\bf Full matrix (baseline)}: standard SwiGLU block, with 3 full matrices and a single activation function on the output of $W_1$;
    \item {\bf Full matrix, double activation}: 3 full matrices, with activations on the outputs of both $W_1$ and $W_3$;
    \item {\bf Low-rank}: parameterization $W_1 = UV$ (no fixed component);
    \item {\bf Low-rank + Identity}: parameterization $W_1= I_{\text{pad}} + UV$, where $I_{\text{pad}}$ is the identity matrix $I_D$ padded at the bottom with zeros;
    \item {\bf Low-rank + Diagonal (LoPA)}: parameterization $W_1= G + UV$, with $[G]_{i,j}=1$ if $j=\left\lfloor i/k \right\rfloor$, and $0$ otherwise, where $k=H/D$;
    \item {\bf Low-rank \;-\; Diagonal (LoPA)}: parameterization $W_1= -G + UV$ (similar to Low-rank + Diagonal, but with $-1$ on the diagonal instead of $+1$)
    \item {\bf Low-rank + Tiled Identity (LoPA)}: parameterization $W_1= T + UV$, where $T$ is the concatenation of identity matrices $I_D$;
    \item {\bf Single activation}: same as Low-rank + Diagonal, but no activation is added to the output of $W_3$.
\end{itemize}
The non-learnable components of the LoPA  variants are visualized in Figure~\ref{fig:fixed-matrices}.
Note that Low-rank + Diag and Low-rank + Tiled Identity are identical up to a permutation. We included their comparison as a sanity check, but their minor differences in NLL are to be attributed to training noise.

\begin{table}[htbp]
    \caption{\textbf{Ablations of different design choices for the $W_1$ matrix.} Unless stated otherwise, $W_1$ is parameterized using \Wparam consisting of a Low-rank + Diagonal matrices, with double activation (activations on the outputs of both $W_1$ and $W_3$). All models are trained with $\silu$ for a duration of $8\Tchin$, without any sparsity loss, and have the same parameter count.
    We report the standard deviation of the NLL of the baseline, computed over 4 seeds. \quad
    \textsuperscript{\dag}Low-rank + Diag and Low-rank + Tiled Identity are identical up to a permutation. We include their comparison as a sanity check. Their minor differences in NLL are to be attributed to training noise.
    }
    \label{tab:w1-design-ablation}
    \centering
    \small
    \setlength{\tabcolsep}{3pt}
    \begin{tabular}{l cccc cccc}
        \toprule
        & \multicolumn{8}{c}{Average validation NLL} \\
        \cmidrule(lr){2-9}
        & \multicolumn{4}{c}{Level 2} & \multicolumn{4}{c}{Level 8} \\
        \cmidrule(lr){2-5} \cmidrule(lr){6-9}
        Parameterization & $r{=}64$ & $r{=}128$ & $r{=}256$ & $r{=}512$ & $r{=}128$ & $r{=}256$ & $r{=}512$ & $r{=}1024$ \\
        \midrule
    Full matrix (baseline) & \multicolumn{4}{c}{$2.800\pm 0.005$} & \multicolumn{4}{c}{$2.152\pm 0.002$} \\
    Full matrix, double activation & \multicolumn{4}{c}{$2.795$} & \multicolumn{4}{c}{$2.153$} \\
    Low-rank & $2.829$ & $2.813$ & $2.804$ & $2.806$ & $2.183$ & $2.167$ & $2.154$ & $2.153$ \\
    Low-rank + Identity & $2.810$ & $2.799$ & $2.795$ & $2.805$ & $2.159$ & $2.156$ & $2.152$ & $2.154$ \\
    Low-rank + Diagonal\textsuperscript{\dag} & $\mathbf{2.787}$ & $\mathbf{2.778}$ & $2.775$ & $\mathbf{2.792}$ & $2.154$ & $\mathbf{2.151}$ & $\mathbf{2.148}$ & $2.156$ \\
    Low-rank \,- Diagonal & $\mathbf{2.787}$ & $2.780$ & $\mathbf{2.774}$ & $2.817$ & $2.157$ & $2.152$ & $2.150$ & $\mathbf{2.152}$ \\
    Low-rank + Tiled Identity\textsuperscript{\dag} & $\mathbf{2.787}$ & $2.779$ & $\mathbf{2.774}$ & $2.793$ & $2.155$ & $\mathbf{2.151}$ & $\mathbf{2.148}$ & $2.158$ \\
    \Wparam w. single activation & $2.804$ & $2.810$ & $2.805$ & $2.811$ & $2.179$ & $2.165$ & $2.155$ & $2.156$ \\
        \bottomrule
    \end{tabular}
\end{table}

In Table~\ref{tab:w1-design-ablation} we show that \Wparam (Low-rank + Diagonal) and Low-rank + Tiled Identity outperform the other variants we tested.
Interestingly, those parameterizations with a $1$ on each row of the fixed matrix are significantly better than Low-rank + Identity, despite representing a space of matrices with the same maximum rank.
Furthermore, while double activation did not significantly improve the NLL in the standard gating setup, it substantially reduced it for the  \Wparam parameterization.

In general, these results justify our choice to use \Wparam as an efficient parameterization of $W_1$.
Furthermore, this experimentally confirms that the position of the $1$s in the row of the fixed matrix $M$ is not meaningful.
This leads us to choose the set of matrices with a single $\pm1$ in each row as the set $\mathcal{C}$ when approximating a full matrix with a low-rank parameterization.

\section{Baselines: additional details}
\label{sec:baselines}
In this section we describe the implementation details of the baselines we report in Tables~\ref{tab:flop-posthoc} and~\ref{tab:flop-trained}.
Sparsity can in principle be induced at several sites of the FFN, each skipping different matmuls: the two inputs ($s_g,s_u$), the gate output $A(W_1x)$ ($s_{\text{pre}}$, the $s$ of Section~\ref{subsec:sparsity-exploitable-flops}), and the gated intermediate $\mathrm{out2}\cdot x_3$ ($s_{\text{gated}}$, which skips further columns of $W_2$). Our method targets the gate, and, with double activation, additionally the gated intermediate; the input sites are what training-free methods such as top-$p$ and TEAL exploit. Table~\ref{tab:ffn-sparsity-sites} catalogs each site and the matmuls it reduces.
\begin{table}[ht]
    \centering
    \caption{
    \textbf{Summary of sparsification approaches.}
    The two FFN sparsity mechanisms and the matmuls each reduces. Input sparsity ($s_g,s_u$; dim $D$) skips input
    coordinates; hidden-activation sparsity ($s_{\mathrm{pre}},s_{\mathrm{gated}}$; dim $H'$)
    skips hidden units.
    }
    \label{tab:ffn-sparsity-sites}
    \begin{tabular}{lllll}
      \toprule
      Symbol & Tensor & Dim & Produced by & Skips \\
      \midrule
      $s_g$               & input $x\!\to\!W_1$      & $D$  & TEAL/top-p            & $W_1$ input read ($V$) \\
      $s_u$               & input $x\!\to\!W_3$      & $D$  & TEAL/top-p            & $W_3$ input read \\
      $s_{\mathrm{pre}}$   & $A(W_1 x)$          & $H'$ & ours (trained), top-p (gate) & $W_3$ cols $+$ $W_2$ rows \\
      $s_{\mathrm{gated}}$ & $\mathrm{out2}\cdot x_3$ & $H'$ & ours, TEAL/top-p      & $W_2$ rows \\
      \bottomrule
    \end{tabular}
  \end{table}

\begin{itemize}
\item \textbf{\relu-fication}~\citep{mirzadeh2024relu}: consists of swapping the FFN activation to \relu during mid-training. No additional $\ell_1$ loss is used so the sparsity cannot be controlled.
\item \textbf{Top-$p$}~\citep{szatkowski2026universal}: a training-free method applied at inference. For an
  activation vector $v$ (per token, over the feature dimension), it keeps the smallest set of
  largest-magnitude entries whose cumulative mass $\sum |v|^{L}$ reaches a fraction greater than $p\in[0, 1]$ of
  the total, and zeros the rest; $p=1$ keeps everything (dense) and smaller $p$ gives more
  sparsity. We use $L=1$ (cumulative absolute value) and choose $p$ to match the
  reference NLL exactly.
  The same $p$ is applied uniformly across all layers. We can target any FFN site independently: the input
  $x$, the gate $A(W_1x)$, the up-projection $W_3x$, the gated intermediate, or the two
  input sites jointly.
  The intermediate site is used for the per-site comparison and all input sites are used for the
  all-FFN comparison.

\item \textbf{TEAL} ~\citep{Liuetal2025}: a training-free method that induces \emph{input}
  activation sparsity by static magnitude thresholding. For each FFN matrix it zeros every
  input entry with $|x|\le t$ before the matrix-vector multiplication, where the threshold $t=t_p$ is a fixed
  scalar per (layer, matrix), calibrated offline as the $p$-quantile of $|x|$. Per-matrix
  sparsities are allocated by the block-wise greedy search (their Algorithm~1) that minimizes
  the block output $\ell_2$ error at a fixed block budget. We calibrate the thresholds on a
  held-out slice of the same evaluation mixture. For comparability with our
  FFN-site sparsity, we apply TEAL to the FFN matrices only---$W_1,W_3$ on the input $x$ and
  $W_2$ on the gated intermediate---leaving attention dense; the reported sparsity is
  therefore FFN-only.
\end{itemize}

\begin{table}[htbp]
  \caption{\textbf{Full version of Table~\ref{tab:flop-posthoc_short}: trained-in sparsity gives a
  larger theoretical FLOP speedup than post-hoc pruning at matched quality.} FFN theoretical
  FLOP speedup, at a shared reference NLL equal to our method at 75\% sparsity.
  Models are pre-trained $4\Tchin$ then cast for the listed mid-training (MT) budget; \cmark{} adds a $1\Tchin$ decay. Table~\ref{tab:flop-posthoc_short} reports the decayed $4\Tchin$ row only; here we give every mid-training budget, with and without decay.
  Casting stays near $2.0\times$ for \Actcasting and $3.2\times$ for \LoPcasting, while top-$p$ and TEAL fall toward $1.3\times$ as the model is mid-trained longer.}
  \centering
  \small
  \begin{tabular}{ll c c ccc}
  \toprule
  & & & & \multicolumn{3}{c}{Theoretical FLOP speedup $\uparrow$} \\
  \cmidrule(lr){5-7}
  FFN Architecture & MT & Decay & Ref. NLL\,$\downarrow$ & Ours @75\% & top-$p$ & TEAL \\
  \midrule
  \multirow{4}{*}{\Actcasting}
   & 1\Tchin & \xmark & 2.326 & \textbf{2.0} & 1.6 & 1.6 \\
   & 2\Tchin & \xmark & 2.307 & \textbf{2.0} & 1.6 & 1.5 \\
   & 4\Tchin & \xmark & 2.294 & \textbf{2.0} & 1.5 & 1.4 \\
   & 4\Tchin & \cmark & 2.154 & \textbf{2.0} & 1.4 & 1.4 \\
  \midrule
  \multirow{4}{*}{\LoPcasting}
   & 1\Tchin & \xmark & 2.331 & \textbf{3.2} & 1.5 & 1.6 \\
   & 2\Tchin & \xmark & 2.309 & \textbf{3.2} & 1.4 & 1.5 \\
   & 4\Tchin & \xmark & 2.294 & \textbf{3.2} & 1.4 & 1.4 \\
   & 4\Tchin & \cmark & 2.153 & \textbf{3.2} & 1.3 & 1.3 \\
  \bottomrule
  \end{tabular}
  \label{tab:flop-posthoc}
\end{table}

\begin{table}[htbp]
  \centering
  \small
    \caption{\textbf{Full version of the \relu-fication comparison of Table~\ref{tab:model-casting-9T}: model casting reaches higher sparsity at a larger theoretical FLOP speedup.}
  Both are trained-activation methods reported at their own operating points.
  Table~\ref{tab:model-casting-9T} reports the decayed $4\Tchin$ operating point only; here we give every mid-training budget, with and without decay.
  At comparable NLL, model casting attains $89\%$ sparsity for both \Wfullrank and \Wparam, whereas \relu-fication plateaus at $75$--$79\%$ and $57$--$62\%$ respectively.}
  \begin{tabular}{ll c ccc ccc}
  \toprule
  & & & \multicolumn{3}{c}{Model casting @90\% (ours)} & \multicolumn{3}{c}{\relu-fication} \\
  \cmidrule(lr){4-6}\cmidrule(lr){7-9}
  FFN Architecture & MT & Decay & NLL\,$\downarrow$ & FLOP sp.\,$\uparrow$ & Spars.\,$\uparrow$ & NLL\,$\downarrow$ & FLOP sp.\,$\uparrow$ & Spars.\,$\uparrow$ \\
  \midrule
  \multirow{4}{*}{\Wfullrank}
   & 1\Tchin & \xmark & 2.340 & \textbf{2.5} & 89\% & 2.330 & 2.0 & 75\% \\
   & 2\Tchin & \xmark & 2.315 & \textbf{2.5} & 89\% & 2.312 & 2.1 & 77\% \\
   & 4\Tchin & \xmark & 2.298 & \textbf{2.5} & 89\% & 2.299 & 2.1 & 79\% \\
   & 4\Tchin & \cmark & 2.160 & \textbf{2.5} & 89\% & 2.159 & 2.0 & 76\% \\
  \midrule
  \multirow{4}{*}{\Wparam}
   & 1\Tchin & \xmark & 2.346 & \textbf{5.8} & 89\% & 2.326 & 2.3 & 62\% \\
   & 2\Tchin & \xmark & 2.323 & \textbf{5.8} & 89\% & 2.304 & 2.3 & 62\% \\
   & 4\Tchin & \xmark & 2.307 & \textbf{5.9} & 90\% & 2.291 & 2.3 & 62\% \\
   & 4\Tchin & \cmark & 2.165 & \textbf{5.8} & 89\% & 2.149 & 2.1 & 57\% \\
  \bottomrule
  \end{tabular}
  \label{tab:flop-trained}
  \end{table}

\section{Case study: Mixture-of-Experts}
\label{sec:moe-case-study}

One advantage of model casting is that it can be used for mid-training a variety of models with different architectures, for example, Mixture-of-Experts (MoE) models.
We train from scratch and successively mid-train with model casting i) the FFN blocks of a MoE model and ii) an alternative MoE router that natively selects active experts leveraging model casting as a case study.

\mypar{Setup.}
The MoE models are initialized from a Llama 3.2 1B dense model, which we convert to a MoE by replicating the original FFN blocks for each expert with some noise added to break symmetries. Each model has eight experts, of which two are active, with no shared expert; pre-training uses a baseline learning rate of $10^{-4}$, a sequence length of $4096$, and a duration of 4\Tchin. We train two variants: one with a cosine schedule (used only for the additional results below) and one with the WSD schedule. After pre-training, we mid-train with model casting and an $\ell_1$ loss for MT durations of 1\Tchin, 2\Tchin and 4\Tchin, comparing against mid-training with \silu as a baseline. MoE is a favorable setting for casting: it reduces the FFN/attention ratio, especially for long sequences, and shrinks the effective batch each expert sees. Experts use a mixed-precision recipe (bf16 non-expert computation, fp32 router, and fp8 expert GEMMs with DeepSeek-V3-style fine-grained scaling); see Table~\ref{tab:mixed-precision} for details.

\mypar{MoE Router with Sigmoid and Top-$k$.}
  Given a token representation $x_t \in \mathbb{R}^{d}$ at position $t$, the router
  computes per-expert logits via a linear projection $\ell_t = W_g x_t \in \mathbb{R}^{E}$,
  where $E$ is the number of routed experts and $W_g \in \mathbb{R}^{E \times d}$ is
  the gating matrix. Expert affinities are obtained by an element-wise
  sigmoid, $s_{t,e} = \sigma(\ell_{t,e}) \in (0,1)$, which treats
  each expert independently and does not force the scores to compete on the simplex (in contrast to the softmax).
  An additive, non-differentiable load-balancing bias $b \in \mathbb{R}^{E}$
  (updated online from per-expert token counts) is added \emph{only for selection},
  and the top-$k$ experts are chosen as
  $\mathcal{T}_t = \operatorname{top\text{-}\!}k\bigl(s_t + b\bigr)$,
  while the routing weights are read from the \emph{unbiased} scores
  $w_{t,e} = s_{t,e}$ for $e \in \mathcal{T}_t$. The weights are optionally
  $L_p$-normalized across the selected experts and rescaled by a global constant
  $\rho$, giving the layer output
  \begin{equation}
      y_t \;=\; \rho \sum_{e \in \mathcal{T}_t} w_{t,e}\, f_e(x_t),
  \end{equation}
  where $f_e(\cdot)$ denotes the $e$-th expert FFN. Because $\sigma(\cdot)$ is
  applied element-wise and the resulting gate directly multiplies the expert
  output, the router weight modulates the contribution of each active expert
  rather than acting purely as a selection signal.

\mypar{MoE Router with \Actcasting.} In addition to applying model casting on MoE FFN blocks, we can replace the typical expert selection mechanism of the MoE router by first using a \silu router together with a sparsifying loss when training from scratch. Subsequently, we apply model casting with the sparsifying loss during mid-training. This mechanism enables us to select which experts are active natively without relying on the top-$k$ mechanism.

In Table~\ref{tab:moe_sparsity_nll}, we present the results of replacing the standard router mechanism of selecting active experts with activation casting. We experimented with setting fixed penalty weights $\lambda \in \{0.1,0.3\}$ or using the adaptive scheme from ReMoE~\citep{wang2025remoe}, setting a target sparsity to match the standard sparsity of the router (two out of eight experts are active) and $\alpha=1.05$. We observe that using activation casting with $\lambda=0$ is the most competitive approach in terms of validation NLL. However, even when we set the target sparsity at 75\% during training, the router sparsity at validation time is about five percentage points below the target.

\begin{table}[htbp]
    \caption{\textbf{MoE's Average NLL and router sparsity}. We group runs by MT budget (4T, 2T, 1T), and rank them by NLL within each group.
    As expected, \RminSplus without sparsity loss on the router results in the best NLL but very little sparsity (the model using all the experts most of the time); best sparsity is obtained using the usual sigmoid and top-$k$ routing, but model casting can find intermediate operating points.
    }
    \centering
    \small
    \begin{tabular}{lcccccc}
        \toprule
        Router & $\lambda$ (PT) & $\lambda$ (MT) & Train & MT & Avg NLL & Avg sparsity \\
        \midrule
        \RminSplus & 0 & 0 & 1T & 4T & \textbf{1.981} & 5\% \\
        \RminSplus, fixed $\lambda$ & 0.1 & 0.1 & 1T & 4T & 1.999 & 49\% \\
        Sigmoid, top-$k$ & --- & --- & 1T & 4T & 2.007 & \textbf{75\%} \\
        \RminSplus, fixed $\lambda$ & 0.3 & 0.1 & 1T & 4T & 2.011 & 65\% \\
        \RminSplus, adaptive $\lambda$ & 0.3 & @70\% & 1T & 4T & 2.023 & 70\% \\
        \RminSplus, adaptive $\lambda$ & 0.1 & @70\% & 1T & 4T & 2.027 & 71\% \\
        \midrule
        \RminSplus & 0 & 0 & 1T & 2T & \textbf{1.997} & 6\% \\
        \RminSplus, fixed $\lambda$ & 0.1 & 0.1 & 1T & 2T & 2.016 & 52\% \\
        Sigmoid, top-$k$ & --- & --- & 1T & 2T & 2.020 & \textbf{75\%} \\
        \RminSplus, fixed $\lambda$ & 0.3 & 0.1 & 1T & 2T & 2.030 & 68\% \\
        \RminSplus, adaptive $\lambda$ & 0.3 & @70\% & 1T & 2T & 2.038 & 70\% \\
        \RminSplus, adaptive $\lambda$ & 0.1 & @70\% & 1T & 2T & 2.039 & 72\% \\
        \midrule
        \RminSplus & 0 & 0 & 1T & 1T & \textbf{2.026} & 8\% \\
        Sigmoid, top-$k$ & --- & --- & 1T & 1T & 2.043 & \textbf{75\%} \\
        \RminSplus, fixed $\lambda$ & 0.1 & 0.1 & 1T & 1T & 2.044 & 53\% \\
        \RminSplus, fixed $\lambda$ & 0.3 & 0.1 & 1T & 1T & 2.062 & 70\% \\
        \RminSplus, adaptive $\lambda$ & 0.1 & @70\% & 1T & 1T & 2.062 & 72\% \\
        \RminSplus, adaptive $\lambda$ & 0.3 & @70\% & 1T & 1T & 2.067 & 69\% \\
        \bottomrule
    \end{tabular}
    \label{tab:moe_sparsity_nll}
\end{table}

\mypar{\Actcasting the MoE FFN blocks.} In contrast to the dense models trained with the WSD learning rate schedule, in Table~\ref{tab:nll_sparsity_cpt_moe}, we observe that using model casting for mid-training a MoE model does not catch up with the pre-trained loss when we mid-train the MoE for 4\Tchin when using \Actcasting in the FFN blocks. In the MoE case, mid-training with \silu catches up with the pre-training final loss when we mid-train for 4\Tchin. 

In Table~\ref{tab:moe_8e_top1_vs_top2}, we report the performance of a MoE with eight experts that selects the top two versus a MoE with eight experts that selects the top one. The latter model is naturally sparser since it has only one active expert. However, this results in a degradation of performance. Hence, generally it is better to have more active experts and use model sparsity to have higher sparsity levels.

\begin{figure}[htbp]
    \centering
    \includegraphics[width=0.8\linewidth]{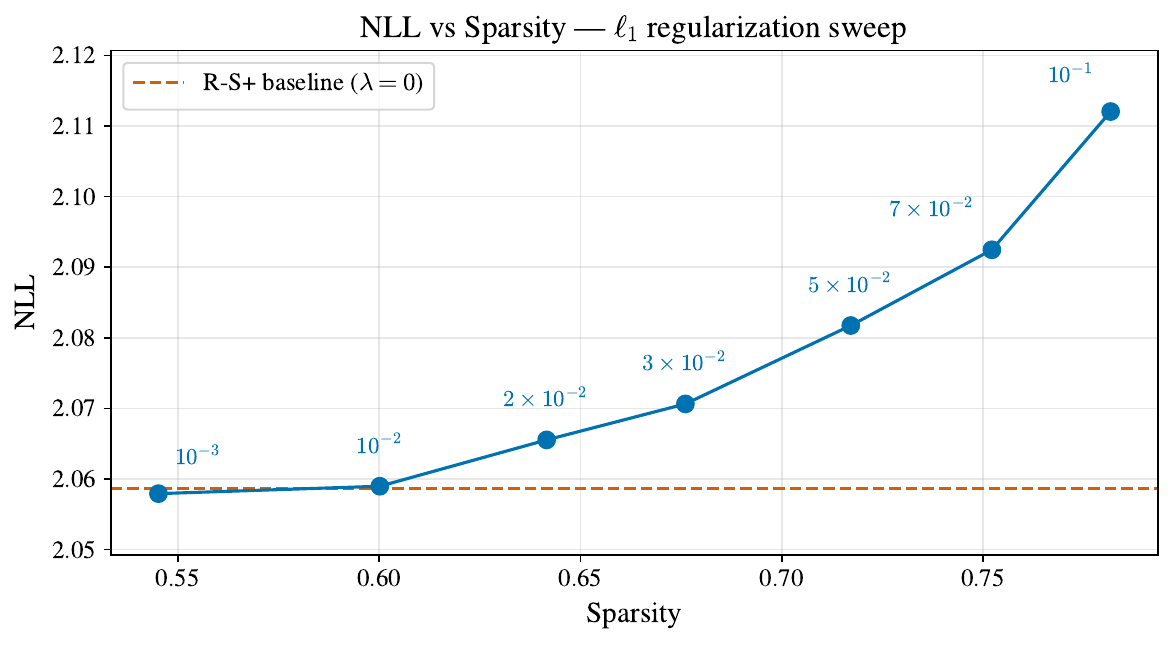}
    \caption{\textbf{MoE FFNs sparsity vs.~NLL} for different values for the $\ell_1$ penalty weight, using cosine schedule. The baseline corresponds to model casting of the FFN to \RminSplus with $\lambda=0$. The sparsity can be increased by about 25\%, but at the cost of considerable NLL degradation, greater than in the non-MoE case.}
    \label{fig:L1_sweep_moe}
\end{figure}

We also explore increasing the penalty weight $\lambda$ to further encourage sparsity in the MoE FFN blocks, in Figure~\ref{fig:L1_sweep_moe}: the sparsity of the MoE can increase by up to 20\%, at the cost of an NLL about $2.6\%$ higher than that of the MoE with $\lambda=0$.

\begin{table}[h]
  \caption{
  \textbf{NLL degradation for MoE FFN \Actcasting.}
  We report the average NLL and sparsity for different penalty weights for $\ell_1$ loss with fixed or adaptive $\lambda$, in the FFN \Actcasting setup for MoEs. Rows with @$x\%$ use an adaptive $\ell_1$ loss for a given sparsity target of $x\%$. We mid-train the MoE with 8 experts top-$2$ with model casting on the experts FFN blocks using a WSD learning rate schedule. All models
  start from the same pre-trained checkpoint, trained for 4\Tchin.}
  \centering
  \begin{tabular}{lccccc}
    \toprule
    Activation & MT & $\lambda$ & NLL & $\Delta$NLL & Sparsity (FFN) \\
    \midrule
    \silu (pre-trained) & 0\Tchin & 0 & 2.008 & +1.36\% & -- \\
    \midrule
    \silu & 4\Tchin & 0 & 1.981 & +0.00\% & -- \\
    \RminSplus & 4\Tchin & 0 & 2.020 & +1.97\% & 56.4\% \\
    \RminSplus & 4\Tchin & $10^{-3}$ & 2.014 & +1.65\% & 59.0\% \\
    \RminSplus & 4\Tchin & $10^{-2}$ & 2.017 & +1.82\% & 66.2\% \\
    \RminSplus & 4\Tchin & @75\% & 2.022 & +2.04\% & 72.8\% \\
    \RminSplus & 4\Tchin & $10^{-1}$ & 2.044 & +3.15\% & 82.4\% \\
    \RminSplus & 4\Tchin & @90\% & 2.064 & +4.19\% & 88.6\% \\
    \bottomrule
  \end{tabular}
  \label{tab:nll_sparsity_cpt_moe}
\end{table}

\begin{table}[h]
  \caption{\textbf{FFN post-activation sparsity vs.~lower number of experts.} Performance of MoE top-$1$ vs MoE top-$2$, out of 8 experts.
  \RminSplus @90\% with 2 active experts results in lower NLL and lower FLOPs than \silu with 1 active expert.
  }
  \centering
  \begin{tabular}{lcccc}
  \toprule
  Activation & top-$k$ & MT & NLL & Sparsity (FFN) \\
  \midrule
  \silu (pre-trained) & 2 & 0\Tchin & 2.008 & -- \\
   \RminSplus @90\% &2& 4\Tchin & 2.064 &  88.6\% \\

  \silu (pre-trained) & 1 & 0\Tchin & 2.091 & -- \\
  \silu & 1 & 4\Tchin & 2.073 & -- \\
  \bottomrule
  \end{tabular}
  \label{tab:moe_8e_top1_vs_top2}
\end{table}

\begin{table}[h!]
    \centering
    \caption{\textbf{Mixed-precision training setup.}
    All fp8 tensors use the E4M3 format (no E5M2).}
    \label{tab:mixed-precision}
    \begin{tabular}{lll}
        \toprule
        \textbf{Component} & \textbf{Precision} & \textbf{FP8 format} \\
        \midrule
        Attention / norms / embeddings / residual            & bf16 & --- \\
        Router GEMM + gating (\silu) + routing losses         & fp32 & --- \\
        Expert FFN --- forward                               & fp8 ($1{\times}128$ act, $128{\times}128$ wt) & E4M3 \\
        Expert FFN --- dgrad (incl.\ grad.\ quant.)          & fp8 (tile-wise $1{\times}128$) & E4M3 \\
        Expert FFN --- wgrad (\silu)                        & fp8 & E4M3 \\
        Expert FFN --- wgrad (\RminSplus) & bf16 & --- \\
        Weight-gradient accumulation                         & fp32 & --- \\
        \bottomrule
    \end{tabular}
\end{table}

\section{NLL vs sparsity for different fixed lambda values}
\label{app:l1_sweep}
\begin{figure}[t]
    \centering
     \includegraphics[width=0.8\linewidth]{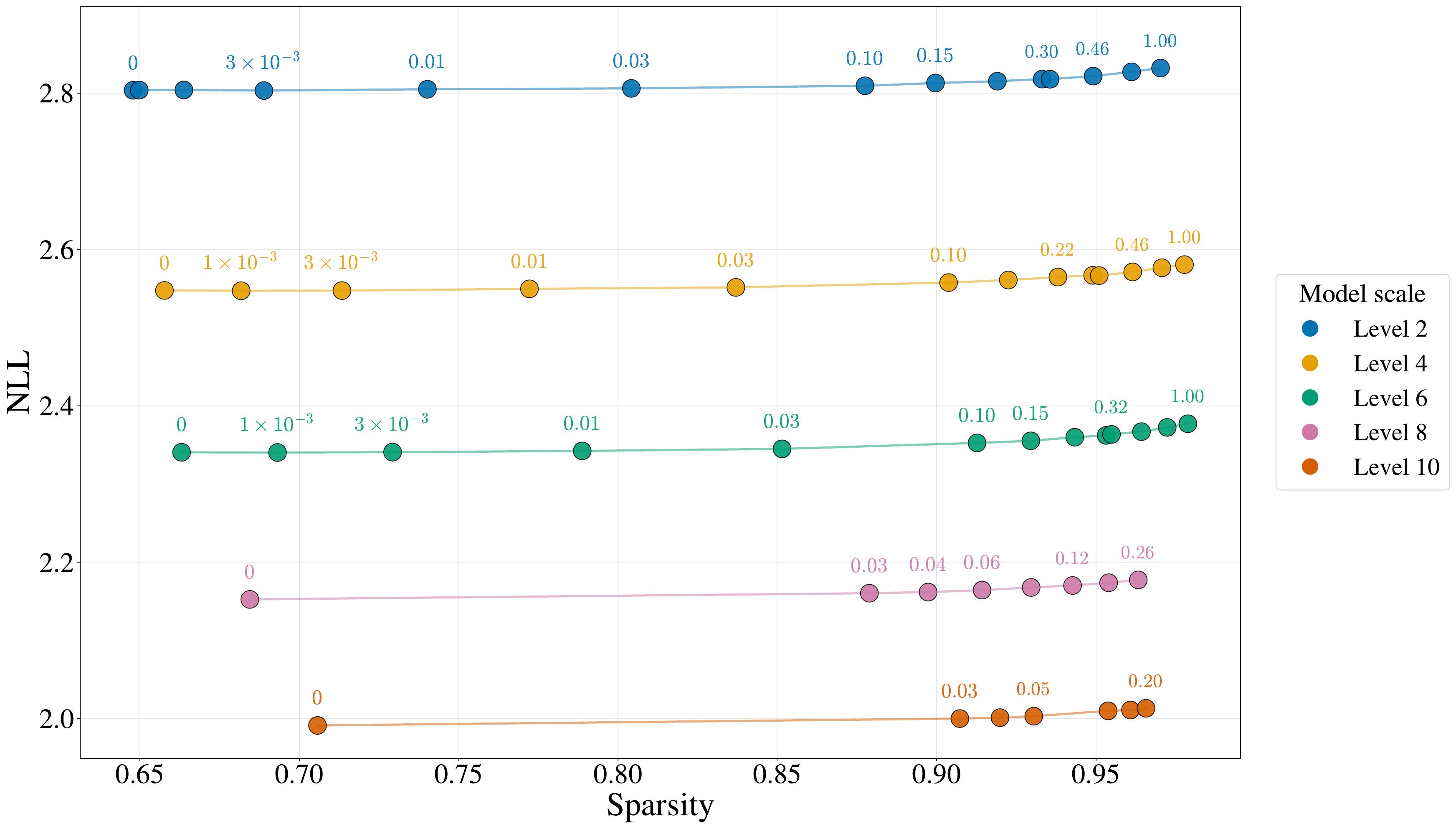}
    \caption{\textbf{Impact of a fixed $\ell_1$ loss.}
    NLL as a function of sparsity for different fixed $\lambda$ values, for level 2, 4, 6, 8 and 10 dense models mid-trained with \Actcasting.
    The numbers on the markers indicate $\lambda$, the weight of the $\ell_1$ loss.
    }
    \label{fig:pareto_with_sparsity_loss}
\end{figure}

Penalizing the gate activation $A(W_1 x)$ with an $\ell_1$ loss during mid-training increases sparsity further without large NLL degradation across a wide range of penalty weights $\lambda$. Figure~\ref{fig:pareto_with_sparsity_loss} shows this sparsity/NLL tradeoff
  for models of levels ranging from 2 to 10: the NLL barely degrades even beyond $90\%$ sparsity, though the sparsity reached at a given $\lambda$ varies with model scale.

Interestingly, we observe that larger models are easier to sparsify.
Without sparsity-inducing loss, the ``natural'' sparsity of the model gradually increases, gaining 6 percentage points when scaling up from level 2 to level 10.
At any fixed value of $\lambda$, the coefficient of the $\ell_1$ loss, the sparsity also increases with respect to the model size.
For instance, at $\lambda=0.03$, the sparsity jumps of 10 percentage points between level 2 and 10, suggesting that larger models are easier to sparsify.

\section{Scaling ladder}
\label{sec:ladder}
We provide in Table~\ref{tab:ladder} the details of the scaling ladder used for all our experiments.

\begin{table}[ht]
  \caption{\textbf{Scaling ladder.} Non-embedding parameters, architecture (FFN input dimension $D$, FFN hidden dimension $H$, number of layers $L$), peak learning rate $\gamma$ and training budget per compute level. The cosine schedule decays to $\gamma/100$.
  \label{tab:ladder}}
  \centering
  \resizebox{\textwidth}{!}{
  \begin{tabular}{
    c
    S[table-format=1.2]
    S[table-format=4.0]
    S[table-format=5.0]
    S[table-format=2.0]
    S[table-format=1.2e1]
    S[table-format=1.1e1]
    S[table-format=4.2]
  }
    \toprule
    & & \multicolumn{3}{c}{Architecture} & & \multicolumn{2}{c}{Training budget} \\
    \cmidrule(lr){3-5} \cmidrule(lr){7-8}
    {Scale} & {Non-Embed Params (B)} & {$D$} & {$H$} & {$L$} & {Peak LR ($\gamma$)} & {Train FLOPs} & {Total Tokens (B)} \\
    \midrule
    0  & 0.03 & 640 & 2048 & 7 & 2.06e-3 & 7.3e18 & 7.65 \\
    1  & 0.05 & 768 & 2048 & 8 & 1.87e-3 & 1.3e19 & 10.73 \\
    2  & 0.07 & 768 & 2048 & 11 & 1.70e-3 & 2.2e19 & 14.86 \\
    3  & 0.12 & 1024 & 3072 & 10 & 1.50e-3 & 5.1e19 & 24.33 \\
    4  & 0.17 & 1024 & 3072 & 14 & 1.35e-3 & 8.9e19 & 33.97 \\
    5  & 0.28 & 1280 & 4096 & 14 & 1.20e-3 & 1.9e20 & 52.85 \\
    6  & 0.37 & 1536 & 4096 & 15 & 1.07e-3 & 3.8e20 & 81.03 \\
    7  & 0.51 & 1536 & 4096 & 21 & 9.67e-4 & 6.9e20 & 113.62 \\
    8  & 0.81 & 1920 & 5120 & 21 & 8.66e-4 & 1.5e21 & 168.44 \\
    9  & 1.26 & 2304 & 6144 & 23 & 7.71e-4 & 3.1e21 & 253.76 \\
    10 & 1.84 & 2560 & 7168 & 26 & 6.98e-4 & 6.1e21 & 359.87 \\
    13 & 5.79 & 3840 & 10240 & 38 & 5.01e-4 & 5.3e22 & 1141.52 \\
    \bottomrule
  \end{tabular}
  }
\end{table}

\section{\Archcasting vs \Wfullrank at reduced parameter count}
\label{sec:arch-cast-vs-reduced}

We compare the \Archcasting models to SwiGLU models trained from scratch with a reduced number of parameters.
This is achieved by reducing the hidden dimension of the model to $H=3712$, so that the number of parameters per FFN is equal to the parameter count of the FFNs after \Archcasting.

The results of this comparison are shown in Table~\ref{tab:arch-cast-vs-reduced}.
The gap between the dense baseline with reduced parameters and the Architecture casting models is initially large, because casting is highly disruptive: the model must simultaneously absorb a new activation and a reduced-rank gate.
However, the gap narrows as the casting time increases, and Architecture casting with \silu eventually outperforms the baseline at iso-parameters.

This shows that \Archcasting produces models with competitive NLL for their parameter count, and that the NLL degradation in Table~\ref{tab:model-casting-9T} is to be attributed to parameter and sparsity differences, not to the architecture change during training.

\begin{table}[t]
\caption{\textbf{\Archcasting produces models with competitive NLL for their parameter count.} Impact of casting time (``MT'' column) on the NLL of models. We compare level 8 models trained with \Archcasting (three first NLL columns) against a dense \silu model trained from scratch with hidden dimension reduced to $H=3712$ so that its FFN parameter count matches that of the cast models.}
\centering
\small
\begin{tabular}{crr|cccc|ccc}
    \toprule
    \multicolumn{3}{c|}{}  & \multicolumn{4}{c|}{Score: NLL}  & \multicolumn{3}{c}{$\Delta$ NLL vs reduced params}  \\
    \multicolumn{3}{c|}{Training duration}  &  \multicolumn{2}{c}{no $\ell_1$}  & $\ell_1$ & Reduced params & \multicolumn{2}{c}{no $\ell_1$} & $\ell_1$  \\
    \cmidrule(l){4-5} \cmidrule(l){8-9}
    PT  &  MT   & Total  &  \silu  & \RminSplus &  \RminSplus & \silu & \silu & \RminSplus & \RminSplus \\
    \midrule
    4\Tchin & 0T & 5\Tchin & 2.240 & 2.255 & 2.276 & \textbf{2.217} & +1.05\% & +1.73\% & +2.66\% \\
    4\Tchin & 0.5T & 5.5\Tchin & 2.216 & 2.226 & 2.240 & \textbf{2.207} & +0.42\% & +0.87\% & +1.53\% \\
    4\Tchin & 1T & 6\Tchin & 2.203 & 2.211 & 2.222 & \textbf{2.200} & +0.16\% & +0.50\% & +1.01\% \\
    4\Tchin & 2T & 7\Tchin & \textbf{2.197} & 2.204 & 2.212 & 2.199 & -0.09\% & +0.22\% & +0.61\% \\
    4\Tchin & 4T & 9\Tchin & \textbf{2.180} & 2.186 & 2.193 & 2.184 & -0.17\% & +0.10\% & +0.40\% \\
    \bottomrule
\end{tabular}
\label{tab:arch-cast-vs-reduced}
\end{table}

\section{Full results for different casting durations}
\label{sec:full-results-cast-duration}
In Table~\ref{tab:model_casting_res}, we present the full results in a setup identical to Table~\ref{tab:model-casting-9T}, across different casting durations.
All models are pre-trained for 4\Tchin, but have a casting phase ranging from 0\Tchin to 4\Tchin; we always finish the casting phase with 1\Tchin tokens of annealing, in which the learning rate is decayed. The results show that the gap between cast models and the \silu reference narrows as the casting period increases.
In particular, \Archcasting presents a wide initial gap due to the large architectural change in the model at casting time, but this gap quickly reduces as well.

\begin{table}[htbp]
    \caption{
    \textbf{Evolution of NLL during casting.}
    Detailed results when training a 0.81B dense model (level 8 in the scaling ladder from Table~\ref{tab:ladder}), performing the three model casting recipes. All models are pre-trained for 4\Tchin; the ``Casting'' column gives the mid-training budget and ``Total'' counts every token spent, including the $1\Tchin$ of decay.
    For LoPA and Architecture Casting, the rank ratio $\alpha$ is set to $1/8$.
    The $\ell_1$ loss targets $90\%$ sparsity at training time; the sparsity measured on the validation set is about one point lower, which is why the $\ell_1$ columns read $89\%$.
    In addition to \texttt{R-S+}, we simulate a model casting with \silu as a control experiment.
    }
    \label{tab:model_casting_res}
    \centering
    \small
    \begin{tabular}{rr|ccc|cc}
        \toprule
        \multicolumn{2}{c|}{Training duration} & \multicolumn{3}{c|}{Score: NLL} & \multicolumn{2}{c}{\RminSplus sparsity} \\
        \cmidrule(lr){1-2} \cmidrule(lr){3-5} \cmidrule(lr){6-7}
        Casting & Total & \silu & \RminSplus & \RminSplus, $\ell_1$ & no $\ell_1$ & $\ell_1$ \\
        \midrule
        \multicolumn{7}{l}{\textit{(a) \Actcasting}} \\
        \quad 0T & 5\Tchin & 2.178 & 2.190 & 2.214 & 61\% & 89\% \\
        \quad 0.5T & 5.5\Tchin & 2.169 & 2.179 & 2.197 & 63\% & 89\% \\
        \quad 1T & 6\Tchin & 2.162 & 2.171 & 2.185 & 65\% & 89\% \\
        \quad 2T & 7\Tchin & 2.161 & 2.170 & 2.179 & 66\% & 89\% \\
        \quad 4T & 9\Tchin & 2.145 & 2.153 & 2.160 & 69\% & 89\% \\
        \addlinespace
        \multicolumn{7}{l}{\textit{(b) \LoPcasting}} \\
        \quad 0T & 5\Tchin & 2.183 & 2.195 & 2.224 & 48\% & 89\% \\
        \quad 0.5T & 5.5\Tchin & 2.176 & 2.181 & 2.205 & 49\% & 89\% \\
        \quad 1T & 6\Tchin & 2.172 & 2.172 & 2.192 & 50\% & 89\% \\
        \quad 2T & 7\Tchin & 2.162 & 2.167 & 2.185 & 51\% & 89\% \\
        \quad 4T & 9\Tchin & 2.150 & 2.148 & 2.165 & 51\% & 89\% \\
        \addlinespace
        \multicolumn{7}{l}{\textit{(c) \Archcasting}} \\
        \quad 0T & 5\Tchin & 2.240 & 2.255 & 2.276 & 69\% & 89\% \\
        \quad 0.5T & 5.5\Tchin & 2.216 & 2.226 & 2.240 & 70\% & 89\% \\
        \quad 1T & 6\Tchin & 2.203 & 2.211 & 2.222 & 71\% & 89\% \\
        \quad 2T & 7\Tchin & 2.197 & 2.204 & 2.212 & 72\% & 89\% \\
        \quad 4T & 9\Tchin & 2.180 & 2.186 & 2.193 & 73\% & 90\% \\
        \bottomrule
    \end{tabular}
\end{table}

\section{Validation NLL for a model with 5.8B weights}
\label{sec:level13-nll}

We validate the applicability of model casting to models of larger scale by applying it to 5.79B-parameter models (level 13 in Table~\ref{tab:ladder}).
We use the same setup as for the level 8 models presented in Table~\ref{tab:model-casting-9T} of the main text and in Table~\ref{tab:model_casting_res} of Appendix~\ref{sec:full-results-cast-duration}.

In Table~\ref{tab:level13-validation-nll} we observe that \LoPcasting is able to achieve a 0.65\% relative degradation in NLL with respect to the dense baseline, a smaller relative gap than for the 0.81B models.
This is consistent with the findings of Figure~\ref{fig:pareto_with_sparsity_loss}: larger models are easier to sparsify, where a fixed $\lambda$ penalty resulted in higher sparsity levels for larger models.

\begin{table}[h]
  \centering
  \small
  \caption{\textbf{Scaling up model casting.}
  Results for 5.79B-parameter models (level 13 in Table~\ref{tab:ladder}).
  $\Delta$NLL is computed relative to the first row in \textbf{bold}, corresponding to a control experiment conducted with \silu throughout training.
  All models are pretrained for 4\Tchin tokens, and midtrained for 4\Tchin tokens of stable learning rate, followed by 1\Tchin tokens of learning rate decay.
  }
  \begin{tabular}{lcccc}
  \toprule
  Setup (pre-training $\rightarrow$ mid-training)
      & NLL & $\Delta$NLL & Sparsity & Speedup \\
  \midrule
  \qquad \Base[\silu{}] $\to$ \Base[\silu{}]~{\small(dense baseline)}
      & \textbf{1.781} & --- & --- & $1.00\times$ \\
  \qquad \LoPA[\silu{}] $\to$
      \LoPA[\RminSplus, $\ell_1$@90\%]
      & 1.792 & $+0.65\%$ & 89\% & $5.86\times$ \\
  \bottomrule
  \end{tabular}
  \label{tab:level13-validation-nll}
\end{table}

\section{Catch up times at mid-training}
\label{sec:catchup}
In Table~\ref{tab:catchup}, we report the catch-up duration across model scales for a fixed 4\Tchin pre-training budget, for both \Actcasting and \LoPcasting.

\begin{table}[htbp]
\caption{
\textbf{Mid-training duration required to catch up} with the \silu model's performance, without the decay phase, for different model sizes with pre-trained models of 4\Tchin duration. Models are evaluated every 0.25\Tchin, and we report the first time at which the finetuned model catches up to the original NLL, rounded up by at most 0.25\Tchin. When the $\ell_1$ loss is applied (rightmost columns), fixing the target sparsity at $90\%$ during training leads to a measured sparsity around $89\%$. For \LoPcasting, the rank ratio $\alpha$ is set to $1/8$.
}
    \label{tab:catchup}
\centering
\footnotesize
\begin{tabular}{cc|ccc|ccc}
    \toprule
    & & \multicolumn{3}{c|}{Casting without $\ell_1$} & \multicolumn{3}{c}{Casting with $\ell_1$} \\
    \cmidrule(lr){3-5}\cmidrule(lr){6-8}
    Model scale & NLL & Duration & NLL & Sparsity & Duration & NLL & Sparsity \\
    \midrule
    \multicolumn{8}{l}{\textit{(a) \Actcasting}} \\
    \quad 2 & 3.013 & 1.00\Tchin & 3.011 & 66\% & 1.50\Tchin & 3.004 & 89\% \\
    \quad 4 & 2.738 & 0.75\Tchin & 2.725 & 67\% & 2.00\Tchin & 2.735 & 89\% \\
    \quad 6 & 2.521 & 1.00\Tchin & 2.518 & 67\% & 1.50\Tchin & 2.516 & 89\% \\
    \quad 8 & 2.321 & 1.25\Tchin & 2.317 & 70\% & 1.75\Tchin & 2.313 & 89\% \\
    \quad 10 & 2.146 & 1.25\Tchin & 2.144 & 71\% & 2.00\Tchin & 2.141 & 89\% \\
    \addlinespace
    \multicolumn{8}{l}{\textit{(b) \LoPcasting}} \\
    \quad 2 & 3.009 & 0.75\Tchin & 3.007 & 48\% & 1.50\Tchin & 3.001 & 89\% \\
    \quad 4 & 2.743 & 0.50\Tchin & 2.742 & 48\% & 2.00\Tchin & 2.739 & 89\% \\
    \quad 6 & 2.523 & 0.75\Tchin & 2.520 & 52\% & 1.75\Tchin & 2.515 & 89\% \\
    \quad 8 & 2.328 & 0.75\Tchin & 2.323 & 56\% & 1.75\Tchin & 2.317 & 89\% \\
    \quad 10 & 2.145 & 1.00\Tchin & 2.141 & 57\% & 2.75\Tchin & 2.137 & 89\% \\
    \bottomrule
\end{tabular}
\end{table}

\section{Quantization-aware training}
\label{sec:qat}

We demonstrate that post-activation sparsity is compatible with quantization by applying Quantization-Aware Training (QAT) on top of models trained with our three model casting recipes.

\mypar{Method.}
We consider 0.81B models (level 8 in Table~\ref{tab:ladder}) pre-trained with \silu for 4T, fine-tuned with model casting for 5T (4T of stable and 1T of decaying learning rate).
We then apply a simple weight quantization approach where weights are converted either to \texttt{int8} or \texttt{int4} at inference. During QAT, we quantize and de-quantize the weights to prepare the model for inference. We apply QAT for 0.5T; optimizer state is reset, a warmup is done for $10\%$ of QAT, and the learning rate then follows a cosine decay.
During QAT, the $\ell_1$ loss is kept to maintain the target sparsity level.
We validate the practical gains of combining quantization and post-activation sparsity by showcasing the performances of low-precision sparse kernels.

\mypar{Results.}
Table~\ref{tab:qat-casting} reports the performance of the different models with and without QAT, at different precisions. For each casting family, we report both the activation swap (\RminSplus + $\ell_1$@90\%), as well as a reference experiment where we use \silu throughout training. Note that Architecture casting underperforms due to its reduced number of parameters.
Interestingly, quantizing sparse models is slightly less costly than dense models. For instance, quantizing to \texttt{int8} the \silu reference of Activation casting costs 0.015 nats, while quantizing the sparse equivalent (\RminSplus + $\ell_1$@90\%) only costs 0.013 nats; the same numbers hold for LoPA casting. Therefore, combining both methods results in sub-additive NLL degradations, suggesting that both methods are independent and compatible.

\begin{figure}[ht]
    \centering
    \includegraphics[width=0.8\linewidth]{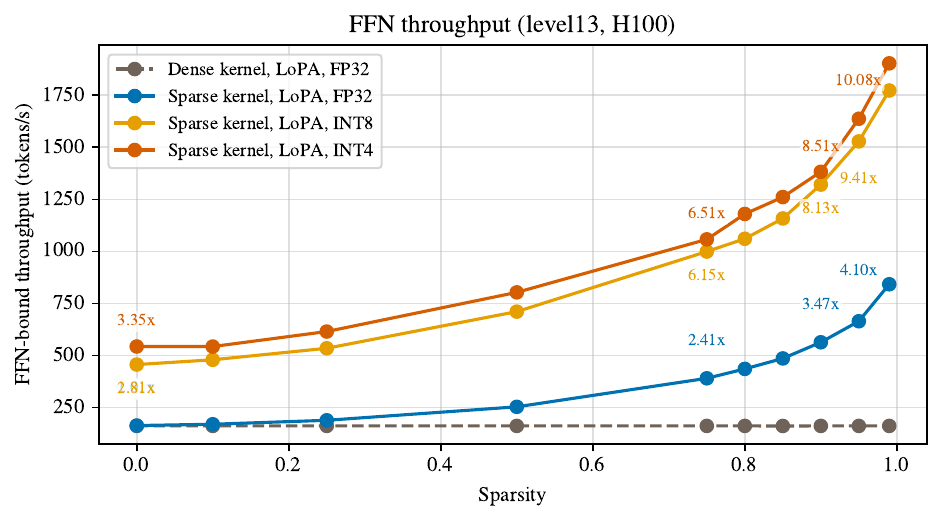}
    \caption{
    \textbf{Measured speedups for quantized sparse models.}
    FFN-bound throughput as a function of the activation sparsity, for a 5.79B non-embedding parameters model (level 13 in the scaling ladder from Table~\ref{tab:ladder}) at different quantization levels (\texttt{int8} and \texttt{int4}).
    }
    \label{fig:qat-kernels}
\end{figure}

Figure~\ref{fig:qat-kernels} demonstrates major speedups when simultaneously exploiting post-activation sparsity and low-precision at inference. Unlike the NLL results above, these kernel measurements are taken on a larger 5.79B model (level 13 in Table~\ref{tab:ladder}). Without sparsity, quantizing the weights to \texttt{int8} and \texttt{int4} respectively provide $2.81\times$ and $3.35\times$ FFN speedups, while LoPA in FP32 results in a $3.31\times$ speedup at 90\% sparsity. Combined together, we obtain speedups of $8.13\times$ (\texttt{int8}) and $8.51\times$ (\texttt{int4}) at 90\% sparsity compared to the full-precision dense baseline.

\begin{table}[ht]
    \caption{
    \textbf{Impact on NLL of quantization and casting.}
    After Activation, LoPA, and Architecture casting, we apply QAT and quantize models in a weight-only \texttt{int4} and \texttt{int8} setting. Degradation is relative to the activation-casting \silu model at full precision in \textbf{bold}.}
    \label{tab:qat-casting}
    \centering
    {\small
    \begin{tabular}{llcccccc}
    \toprule
     & & \multicolumn{2}{c}{FP32} & \multicolumn{2}{c}{\texttt{int8}} & \multicolumn{2}{c}{\texttt{int4}} \\
     \cmidrule(lr){3-4} \cmidrule(lr){5-6} \cmidrule(lr){7-8}
    \multicolumn{1}{c}{Casting family} & \multicolumn{1}{c}{Variant} & NLL & $\Delta$NLL (\%) & NLL & $\Delta$NLL (\%) & NLL & $\Delta$NLL (\%) \\
    \midrule
    \multirow{2}{*}{Activation} & \silu & \textbf{2.145} & --- & 2.160 & +0.70\% & 2.178 & +1.54\% \\
     & \RminSplus + $\ell_1$@90\% & 2.160 & +0.68\% & 2.173 & +1.29\% & 2.188 & +2.01\% \\
    \midrule
    \multirow{2}{*}{LoPA} & \silu & 2.147 & +0.09\% & 2.162 & +0.76\% & 2.180 & +1.62\% \\
     & \RminSplus + $\ell_1$@90\% & 2.166 & +0.99\% & 2.179 & +1.57\% & 2.194 & +2.29\% \\
    \midrule
    \multirow{2}{*}{Architecture} & \silu & 2.180 & +1.63\% & 2.193 & +2.21\% & 2.213 & +3.16\% \\
     & \RminSplus + $\ell_1$@90\% & 2.193 & +2.21\% & 2.205 & +2.76\% & 2.222 & +3.59\% \\
    \bottomrule
    \end{tabular}
    }
\end{table}

\section{Enforcing sparsity with \texorpdfstring{$\ell_1$}{L1} loss only}
\label{sec:only_l1}
\begin{figure}[htbp]
      \centering
      \begin{subfigure}[t]{0.48\linewidth}
          \includegraphics[width=\linewidth]{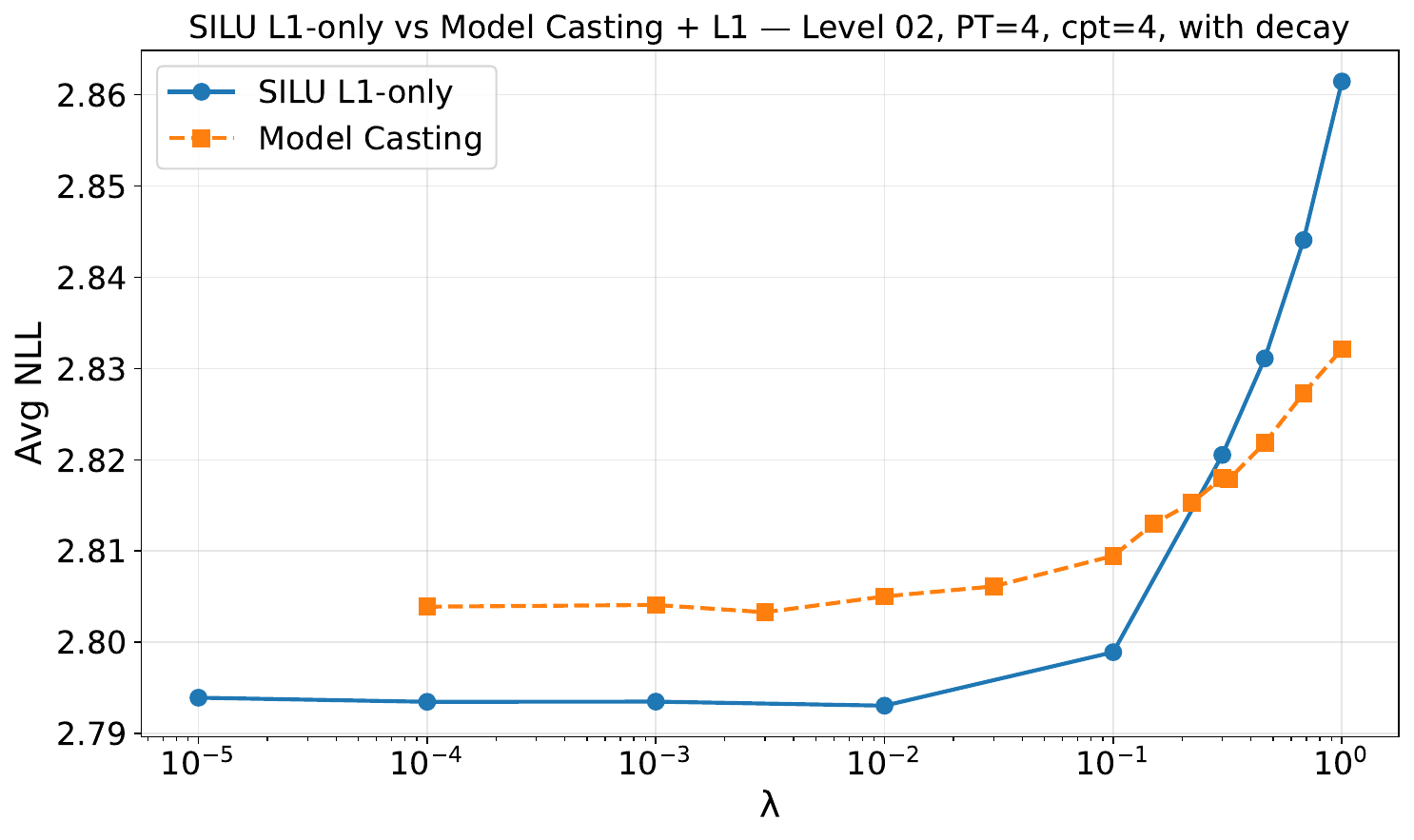}
          \caption{NLL vs sparsity penalty for model casting +$\ell_1$ vs $\ell_1$ only.}
      \end{subfigure}
      \hfill
      \begin{subfigure}[t]{0.48\linewidth}
          \includegraphics[width=\linewidth]{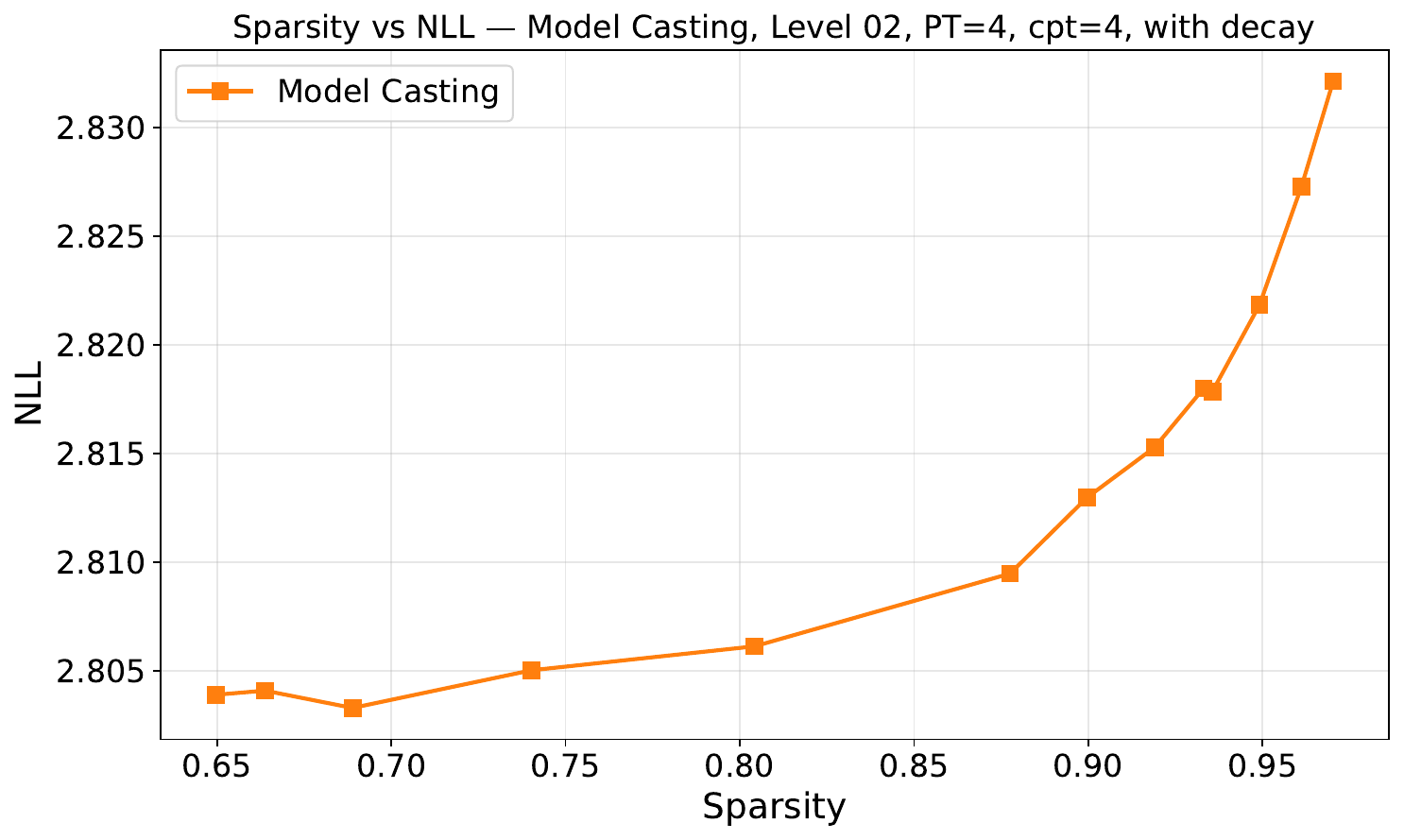}
          \caption{Sparsity vs NLL with model casting + $\ell_1$}
      \end{subfigure}

      \vspace{1em}

      \begin{subfigure}[t]{0.48\linewidth}
          \includegraphics[width=\linewidth]{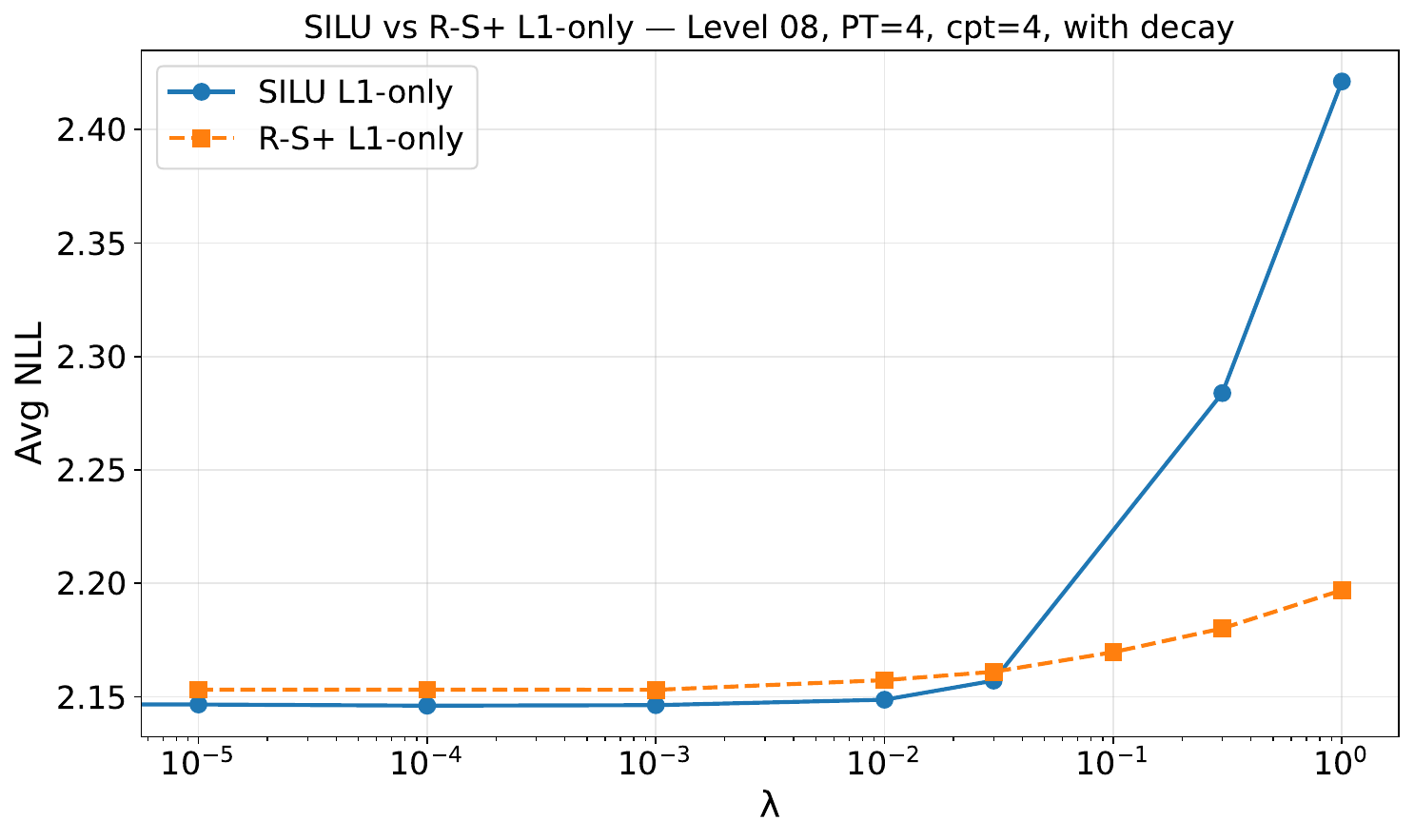}
          \caption{NLL vs sparsity penalty for model casting +$\ell_1$ vs $\ell_1$ only.}
      \end{subfigure}
      \hfill
      \begin{subfigure}[t]{0.48\linewidth}
          \includegraphics[width=\linewidth]{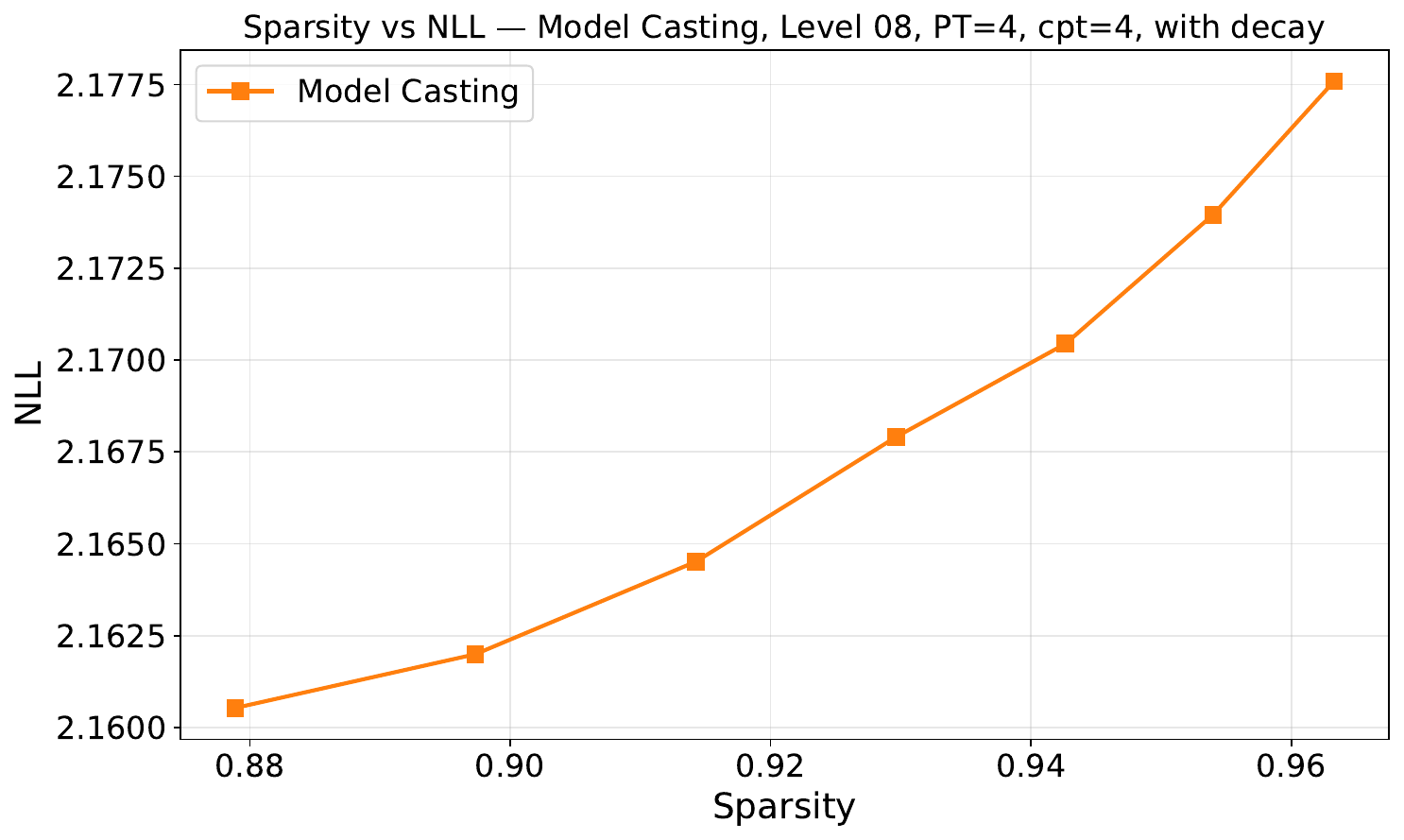}
          \caption{Sparsity vs NLL with model casting + $\ell_1$}
      \end{subfigure}
      \caption{$\ell_1$ loss without model casting induces higher performance degradation for the same sparsity penalty $\lambda$ with level 2 (top) and level 8 (bottom) models.}
      \label{fig:model_casting_l1_wins}
  \end{figure}

Figure~\ref{fig:model_casting_l1_wins} shows that enforcing sparsity by only adding an $\ell_1$ loss to the cross entropy loss, a) incurs a higher performance degradation at a given sparsity penalty, and b) does not result in significant sparsity levels.

\section{Ablating the use of \texorpdfstring{$\ell_1$}{L1} loss during pre-training}
\label{sec:abl-l1-pt}
In this section, we ask whether applying an $\ell_1$ loss during pre-training with \silu, before the model casting phase, could be beneficial.
To do so, we train models from scratch for 4\Tchin using an $\ell_1$ penalty loss on the output of \silu; we then apply activation casting for 5\Tchin tokens (4\Tchin of stable learning rate schedule, and an annealing phase of 1\Tchin), using \RminSplus and $\ell_1$ loss with the same weight as during pre-training.

Table~\ref{tab:lora-pretraining-ablation} shows that adding the $\ell_1$ loss during pre-training results in similar or worse NLL (up to training noise), and reduces the sparsity level after casting.
Furthermore, using $\ell_1$ loss during pre-training prevents the use of an off-the-shelf model, as it requires a different training recipe from scratch.
Therefore, these results indicate that applying model casting as a mid-training step only is sufficient.

\begin{table}[ht]
    \caption{
    \textbf{Using $\ell_1$ loss during pre-training degrades final sparsity.}
    Ablations of using the $\ell_1$ loss during pre-training. Models are pre-trained and cast with the same $\ell_1$ weight $\lambda$.
    Adding the $\ell_1$ loss on the output of \silu during pre-training results in similar or worse NLL (up to training noise), and reduces the sparsity level after casting.}
    \label{tab:lora-pretraining-ablation}
    \centering
    \small
    \setlength{\tabcolsep}{4pt}
    \begin{tabular}{lc|cccccc}
        \toprule
        & & \multicolumn{6}{c}{$\ell_1$ loss weight ($\lambda$)} \\
        \cmidrule(lr){3-8}
        Metric & $\ell_1$ during PT & $10^{-4}$ & $3\times10^{-4}$ & $10^{-3}$ & $3\times10^{-3}$ & $10^{-2}$ & $3\times10^{-2}$ \\
        \midrule
        \multicolumn{8}{l}{\textit{Level 2}} \\
        \quad NLL & \ding{51} & $\mathbf{2.749}$ & $\mathbf{2.748}$ & $\mathbf{2.747}$ & $2.750$ & $2.751$ & $2.754$ \\
        \quad NLL & \ding{55} & $2.750$ & $2.750$ & $2.750$ & $\mathbf{2.749}$ & $\mathbf{2.750}$ & $\mathbf{2.753}$ \\
        \cmidrule(lr){1-8}
        \quad Sparsity & \ding{51} & $\mathbf{65.3\%}$ & $65.6\%$ & $66.7\%$ & $68.1\%$ & $71.0\%$ & $73.6\%$ \\
        \quad Sparsity & \ding{55} & $65.2\%$ & $\mathbf{65.7\%}$ & $\mathbf{66.8\%}$ & $\mathbf{69.3\%}$ & $\mathbf{74.5\%}$ & $\mathbf{80.9\%}$ \\
        \addlinespace
        \midrule
        \multicolumn{8}{l}{\textit{Level 8}} \\
        \quad NLL & \ding{51} & $\mathbf{2.159}$ & $\mathbf{2.160}$ & $\mathbf{2.159}$ & $\mathbf{2.160}$ & $2.164$ & $2.173$ \\
        \quad NLL & \ding{55} & $2.160$ & $2.161$ & $2.162$ & $2.162$ & $\mathbf{2.163}$ & $\mathbf{2.166}$ \\
        \cmidrule(lr){1-8}
        \quad Sparsity & \ding{51} & $\mathbf{73.6\%}$ & $\mathbf{74.1\%}$ & $74.5\%$ & $78.5\%$ & $\mathbf{83.8\%}$ & $88.3\%$ \\
        \quad Sparsity & \ding{55} & $\mathbf{73.6\%}$ & $\mathbf{74.1\%}$ & $\mathbf{76.6\%}$ & $\mathbf{80.0\%}$ & $\mathbf{83.8\%}$ & $\mathbf{88.7\%}$ \\
        \bottomrule
    \end{tabular}
\end{table}

\section{Casting with other activation functions}
\label{sec:diff_activations}

Model casting only requires that the replacement activation produce exact zeros; \RminSplus is
one choice among many. Here we cast the same pre-trained models with a range of alternative
activations and report the resulting NLL and FFN activation sparsity in
Table~\ref{tab:nll_diff_activations}, for a level 2 and a level 8 model, across mid-training
budgets of $1\Tchin$, $2\Tchin$ and $4\Tchin$, with and without the final decay.
All runs share the pre-trained checkpoint and differ only in the activation used at casting time.

The comparison separates two properties that are easily conflated. \silu, G-S+ and \gelu give the
best NLL, but they are not sparsifying activations at all, so they bound the attainable quality
rather than competing as candidates. Among the sparsifying activations, Shifted \relu reaches the
highest sparsity ($75$--$79\%$) but pays for it in NLL, and the $p$-\relu family trades sparsity
away as $p$ grows, from $66\%$ at $p=1.189$ down to $51\%$ at $p=1.681$. \RminSplus sits at the
useful point of that trade-off: its NLL tracks \silu to within a few thousandths at every budget,
while still delivering $65$--$69\%$ sparsity before any $\ell_1$ penalty is applied. Notably
\srelu, which \citet{Zhang2024ReLU2WD} advocate for from-scratch training, gives the
\emph{lowest} exact-zero sparsity here ($40\%$), which is why we do not adopt it for casting.

\begin{table}[htbp]
  \caption{
  \textbf{\silu and \gelu result in the best performances for casting.}
  NLL and FFN activation sparsity when casting with different activation functions.
    Both models are pre-trained for $4\Tchin$; each NLL column is a mid-training budget, and
    ``+D'' adds the final $1\Tchin$ decay. Sparsity varies little with the budget, so we report
    it at the decayed checkpoint only. \silu, G-S+ and \gelu are not sparsifying activations and
    have no exact zeros to report; they bound the attainable NLL rather than competing on
    sparsity. Bold marks the best NLL per column.}
  \label{tab:nll_diff_activations}
  \centering
  \small
  \setlength{\tabcolsep}{4pt}
  \begin{tabular}{l cccc c cccc c}
    \toprule
    & \multicolumn{5}{c}{Level 2} & \multicolumn{5}{c}{Level 8} \\
    \cmidrule(lr){2-6} \cmidrule(lr){7-11}
    & \multicolumn{4}{c}{NLL\,$\downarrow$} & Spars. & \multicolumn{4}{c}{NLL\,$\downarrow$} & Spars. \\
    \cmidrule(lr){2-5} \cmidrule(lr){7-10}
    Activation & $1\Tchin$ & $2\Tchin$ & $4\Tchin$ & +D & +D & $1\Tchin$ & $2\Tchin$ & $4\Tchin$ & +D & +D \\
    \midrule
    \RminSplus & 3.011 & 2.984 & 2.966 & 2.804 & 65\% & 2.326 & 2.307 & 2.295 & 2.153 & 69\% \\
    \silu & {\bf 2.999} & {\bf 2.970} & 2.945 & {\bf 2.794} & --- & 2.326 & 2.307 & 2.295 & 2.153 & --- \\
    \texttt{G-S+} & {\bf  2.999} & 2.974 & {\bf 2.944} & 2.808 & --- & {\bf 2.314} & 2.298 & {\bf 2.287} & 2.148 & --- \\
    \gelu & {\bf  2.999} & 2.973 & 2.955 & 2.808 & --- & {\bf 2.314} & {\bf 2.297} & {\bf 2.287} & {\bf 2.147} & --- \\
    \prelu ($p{=}1.189$) & 3.022 & 2.999 & 2.956 & 2.811 & 66\% & 2.327 & 2.309 & 2.295 & 2.154 & 70\% \\
    \prelu ($p{=}1.414$) & 3.012 & 2.983 & 2.954 & 2.804 & 59\% & 2.325 & 2.306 & 2.293 & 2.152 & 62\% \\
    \prelu ($p{=}1.681$) & 3.013 & 2.984 & 2.966 & 2.803 & 51\% & 2.326 & 2.306 & 2.292 & 2.151 & 52\% \\
    \srelu & 3.014 & 2.989 & 2.966 & 2.803 & 40\% & 2.327 & 2.306 & 2.291 & 2.151 & 40\% \\
    $0.5\,$\relu & 3.022 & 2.999 & 2.960 & 2.811 & 71\% & 2.330 & 2.311 & 2.298 & 2.158 & 75\% \\
    \relu & 3.022 & 2.990 & 2.961 & 2.811 & 72\% & 2.330 & 2.312 & 2.299 & 2.159 & 76\% \\
    Shifted \relu ($\delta{=}1.2$) & 3.020 & 2.991 & 2.960 & 2.811 & 75\% & 2.334 & 2.313 & 2.301 & 2.160 & 79\% \\
    \bottomrule
  \end{tabular}
\end{table}

\section{Qwen3 mid-training details}
\label{sec:qwen3}

We apply our existing mid-training recipe, in terms of the scaling ladder Chinchilla units ($\Tchin=10\,400$, for the model that is closest in size, level 8, which has 0.81B non-embedding parameters, this corresponds to 20 TPP) as follows:
we mid-train for 4\Tchin of warmup-stable (from which \Tchin/5 is for warmup) and then with decay for 1\Tchin (this corresponds to mid-training for 100 TPP).

\mypar{Short sweep for determining the learning rate for mid-training.} We conduct a sweep to determine which learning rate least degrades the validation NLL; the results of this sweep are presented in Tables~\ref{tab:qwen3-lr-sweep} and~\ref{tab:qwen3-lr-sweep-nll}, respectively for short context evals and validation NLL.

\mypar{Qwen3 mid-training with $\gamma=7.4\times10^{-5}$.} Validation NLL is reported in Table~\ref{tab:qwen3-full-nll}; the corresponding short-context evaluations are in Table~\ref{tab:qwen3-delta-5T} of Appendix~\ref{app:short_context}.

\mypar{Validation NLL.} In Tables~\ref{tab:qwen3-lr-sweep-nll} and~\ref{tab:qwen3-full-nll} we report NLL with respect to different data mixtures: \emph{In-mixture} covers five datasets included in the training data mixture; \emph{All 9} averages
nine evaluation  datasets; \emph{GovReport} is an out-of-mixture
probe, reported separately but included in \emph{All 9}.
The nine therefore comprise the five in-mixture datasets and four out-of-mixture probes, of which
\emph{GovReport} is the one we single out.

\begin{table}[htbp]
    \caption{\textbf{Short context evaluations for different peak-lr for mid-training Qwen3-1.7B.} The evaluation is computed
      post-cooldown at step 1{,}250. All arms branch
      from \texttt{Qwen3-1.7B-Base} with the \texttt{R-S+AO} activation swap. Bold marks the best arm
      per task; $^{\ast}$ marks the chosen LR. }
    \centering
    \small
    \setlength{\tabcolsep}{5pt}
    \begin{tabular}{l cccccc}
      \toprule
      & \multicolumn{6}{c}{Peak LR $\gamma$} \\
      \cmidrule(lr){2-7}
      Task & $1.00\times 10^{-5}$ & $7.40\times 10^{-5}$\,\textsuperscript{$\ast$} & $1.05\times 10^{-4}$ & $1.48\times 10^{-4}$ & $2.50\times 10^{-4}$ & $7.40\times 10^{-4}$ \\
      \midrule
      \multicolumn{7}{l}{\textit{Core-12 short-context benchmarks}} \\
      HSwag & .6279 & .6451 & \bf{ .6458} & .6439 & .6359 & .5969 \\
      ARC-e & .7268 & .7281 & \bf{.7328} & .7230 & .7087 & .6283 \\
      ARC-c & .4489 & .4575 & \bf{ .4592} & .4403 & .4206 & .3519 \\
      PIQA & .7361 & \bf{.7503} & .7470 & .7448 & \bf{.7503} & .7388 \\
      WinoG & .6219 & \bf{.6306} & .6267 & \bf{.6306} & .6235 & .6140 \\
      OBQA & .3820 & .3980 & .4000 & \bf{.4020} & .3780 & .3680 \\
      RACE-m & \bf{.6121} & .6100 & .6038 & .5968 & .5815 & .5306 \\
      RACE-h & \bf{.4437} & .4342 & .4234 & .4185 & .4177 & .3848 \\
      HE+ & \bf{.2988} & .2744 & .2927 & .2866 & .2805 & .1646 \\
      MBPP & .4020 & .4200 & \bf{.4320} & .4160 & .3920 & .2420 \\
      TQA & .3207 & .3401 & \bf{.3402} & .3352 & .3175 & .2544 \\
      NQ & .1527 & .1571 & \bf{.1580} & .1572 & .1542 & .1229 \\
      \cmidrule(lr){1-7}
      \textbf{Mean} & .4811 & .4871 & \bf{.4885} & .4829 & .4717 & .4164 \\
      \midrule
      \multicolumn{7}{l}{\textit{Knowledge / commonsense}} \\
      MMLU & .5749 & \bf{.5920} & .5904 & .5841 & .5619 & .4245 \\
      CSQA & .6830 & \bf{.6880} & .6749 & .6806 & .6585 & .5389 \\
      BoolQ & \bf{.6911} & .6749 & .6719 & .6725 & .6685 & .6443 \\
      \bottomrule
    \end{tabular}
    \label{tab:qwen3-lr-sweep}
  \end{table}

 \begin{table}[htbp]
     \caption{
     \textbf{Validation NLLs using different peak-lr for mid-training Qwen3-1.7B.}
      Held-out NLL per token and FFN activation sparsity for Qwen3-1.7B.
      Every arm runs $0.093T$ (1{,}250 steps, 2.62\,B tokens) without $\ell_1$ loss.
      Bold marks the best arm per NLL column; $^{\ast}$ marks the chosen learning rate.}
    \centering
    \small
    \begin{tabular}{l cc c c c}
      \toprule
      Peak LR $\gamma$ & In-mixture & $\Delta$ vs.\ base & All 9 & GovReport & Sparsity (\%) \\
      \midrule
      Pretrained \textsc{Base} & 1.711 & --- & 1.843 & 2.152 & --- \\
      Swap, 0 steps & 10.256 & $+8.544$ & 9.996 & 9.788 & 76.77 \\
      \midrule
      $1.00\times 10^{-5}$ & 1.761 & $+0.050$ & 1.927 & 2.212 & 79.79 \\
      $7.40\times 10^{-5}$\,\textsuperscript{$\ast$} & \textbf{1.724} & $\mathbf{+0.013}$ & \textbf{1.895} & \textbf{2.188} & 78.59 \\
      $1.05\times 10^{-4}$ & 1.727 & $+0.015$ & 1.898 & 2.190 & 78.60 \\
      $1.48\times 10^{-4}$ & 1.734 & $+0.022$ & 1.906 & 2.195 & 78.76 \\
      $2.50\times 10^{-4}$ & 1.757 & $+0.046$ & 1.932 & 2.213 & 79.40 \\
      $7.40\times 10^{-4}$ & 1.871 & $+0.159$ & 2.047 & 2.298 & 82.96 \\
      \bottomrule
    \end{tabular}
    \label{tab:qwen3-lr-sweep-nll}
  \end{table}

\begin{table}[htbp]
    \centering
        \caption{
        \textbf{Mid-training Qwen3-1.7B with model casting.}
        Held-out NLL per token and FFN activation sparsity for
      Qwen3-1.7B mid-training at the swept-optimum peak $\gamma=7.40\times10^{-5}$.
      Runs differ only in FFN activation and sparsity penalty: \RminSplus is
      the swap with no penalty, ``\Actcasting @90\%'' adds an adaptive-$\ell_1$ penalty
      targeting 90\% sparsity, and \emph{\silu, no swap} is plain mid-training.
      Bold marks the best trained
      checkpoint per NLL column.
      }
    \small

      \centering
      \small
      \setlength{\tabcolsep}{4pt}
      \begin{tabular}{l r cc c c c}
        \toprule
        Run & Budget & In-mixture & $\Delta$ vs.\ base & All 9 & GovReport & Sparsity (\%) \\
         &  &  &  & &  & (All 9) \\
        \midrule
        Pretrained \textsc{Base} & --- & 1.711 & --- & 1.843 & 2.152 & --- \\
        Swap, 0 steps & $0T$ & 10.256 & $+8.544$ & 9.996 & 9.788 & 76.77 \\
        \midrule
        \RminSplus & $1T$ & 1.702 & $-0.010$ & 1.881 & 2.163 & 76.84 \\
         & $2T$ & 1.698 & $-0.013$ & 1.881 & 2.155 & 76.43 \\
         & $3T$ & 1.699 & $-0.012$ & 1.884 & 2.151 & 76.25 \\
         & $4T$ & 1.701 & $-0.011$ & 1.887 & 2.151 & 76.11 \\
         & $5T^{\dagger}$ & 1.665 & $-0.046$ & 1.854 & 2.116 & 75.18 \\
        \midrule
        \Actcasting @90\% & $1T$ & 1.730 & $+0.019$ & 1.912 & 2.190 & 89.57 \\
         & $2T$ & 1.722 & $+0.010$ & 1.908 & 2.179 & 89.62 \\
         & $3T$ & 1.719 & $+0.008$ & 1.906 & 2.172 & 89.57 \\
         & $4T$ & 1.719 & $+0.007$ & 1.907 & 2.168 & 89.60 \\
         & $5T^{\dagger}$ & 1.685 & $-0.027$ & 1.875 & 2.136 & 89.54 \\
        \midrule
        \silu, no swap & $1T$ & 1.689 & $-0.023$ & 1.868 & 2.155 & --- \\
         & $2T$ & 1.690 & $-0.022$ & 1.873 & 2.150 & --- \\
         & $3T$ & 1.692 & $-0.019$ & 1.878 & 2.147 & --- \\
         & $4T$ & 1.695 & $-0.017$ & 1.881 & 2.146 & --- \\
         & $5T^{\dagger}$ & \textbf{1.660} & $\mathbf{-0.051}$ & \textbf{1.849} & \textbf{2.113} & --- \\
        \bottomrule
      \end{tabular}
    \label{tab:qwen3-full-nll}
  \end{table}

\section{GPU speedup in bfloat16}
\label{sec:timing-bf16}

\begin{figure}[H]
    \centering
    \includegraphics[width=.85\linewidth]{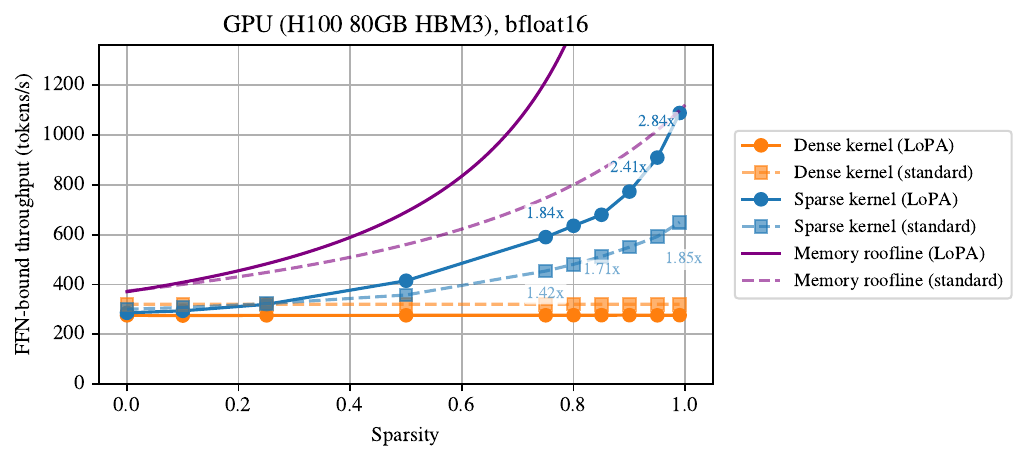}
    \caption{
    \textbf{Measured speedups in bfloat16.}
    Throughput as a function of the activation sparsity, for a 5.79B non-embedding parameters model (level 13 in the scaling ladder from Table~\ref{tab:ladder}) in bfloat16.
    At 90\% sparsity the speedup on GPU is $1.71\times$ for \Wfullrank and $2.41\times$ for \Wparam compared to the standard dense case.
    }
    \label{fig:runtimes-bfloat16}
\end{figure}

As inference is frequently done in bfloat16, we also provide in Figure~\ref{fig:runtimes-bfloat16} the run times observed for this format.

\section{Short-context evaluations}
\label{app:short_context}
In this section, we provide short-context evaluations to complement the validation NLL results presented in Table~\ref{tab:model_casting_res}. We employ two types of benchmarks for zero or few-shot evaluation, which we describe herein, see Table~\ref{tab:benchmarksconfigs} for a summary. The first type is code generation tasks: HumanEval+~\citep{evalplus} and  MBPP~\citep{Austin2021ProgramSW}.  
The second type consists of commonsense and general reasoning: HellaSWAG~\citep{Zellers2019HellaSwagCA}, ARC~\citep{Clark2018ThinkYH}, PIQA~\citep{Bisk_Zellers_Lebras_Gao_Choi_2020}, OBQA~\citep{mihaylov-etal-2018-suit}, WinoGrande~\citep{Sakaguchi_LeBras_Bhagavatula_Choi_2020}, NQ~\citep{kwiatkowski-etal-2019-natural}, RACE~\citep{Lai2017RACELR} and TQA~\citep{TQA2017}.
For the Qwen3 experiments of Appendix~\ref{sec:qwen3} we additionally report three knowledge and commonsense benchmarks: MMLU~\citep{hendrycks2021mmlu}, CSQA~\citep{talmor-etal-2019-commonsenseqa} and BoolQ~\citep{clark-etal-2019-boolq}.
For the post-trained models of Appendix~\ref{sec:post-training} we also report instruction-following with IFEval~\citep{zhou2023ifeval}.

\mypar{Code generation}
We use two benchmarks that evaluate the code generation capabilities of AI models: HumanEval+ and MBPP.
\begin{itemize}
    \item The HumanEval+~\citep{evalplus} benchmark is an extension of HumanEval~\citep{chen2021codex}, which is designed to evaluate the functional correctness of code generated by AI models.
    \item MBPP~\citep{Austin2021ProgramSW} is designed to evaluate the code generation abilities of AI models, particularly for Python programming tasks.
\end{itemize}
\begin{table}[t]
    \caption{\textbf{Summary of benchmarks used for evaluation.} We prompt level 8 models in a few-shot setting, and report a metric that depends on the benchmark, as well as the average performance.
    \texttt{acc\_char} is the Character-Length Normalized Accuracy.
    \label{tab:benchmarksconfigs}}
\centering
{\small
\begin{tabular}{llcc}
\toprule
\textbf{Benchmark} & \textbf{Metric} & \textbf{\#~Shots} & \textbf{Type} \\
\midrule
HellaSwag        & \texttt{acc\_char}      &     0              &    Choice          \\
WinoGrande       & \texttt{acc\_char}      &      0             &       Choice       \\
ARC-e        & \texttt{acc\_char}      &       0            &       Choice        \\
ARC-c   & \texttt{acc\_char}      &      0             &      Choice        \\
PIQA             & \texttt{acc\_char}      &       0            &    Choice         \\
OBQA             & \texttt{acc\_char}      &      0             &          Choice    \\
RACE-m      & \texttt{acc\_char}      &        0           &     Choice         \\
RACE-h        & \texttt{acc\_char}      &       0            &     Choice         \\
HumanEval+ & pass@1     &     0              &     Generation         \\
MBPP             & pass@1         &      3             &        Generation      \\
TriviaQA              & f1             &       5            &      Generation        \\
NaturalQ               & f1             &         5          &   Generation           \\
\bottomrule
\end{tabular}}
\end{table}

\mypar{Common sense and general reasoning}  We use benchmarks consisting of question-answer or multiple-choice questions designed to evaluate the commonsense reasoning abilities of AI models, particularly in the context of natural language understanding: HellaSWAG, ARC, PIQA, OBQA, Winogrande, NaturalQuestions, RACE and TQA.

\begin{itemize}
    \item HellaSwag~\citep{Zellers2019HellaSwagCA} consists of multiple-choice questions where each question contains a short context (a sentence or paragraph) followed by four possible continuations. Only one continuation is correct and makes sense given the context.
    \item The AI2 Reasoning Challenge (ARC)~\citep{Clark2018ThinkYH} benchmark consists of multiple-choice science questions typically found in elementary and middle school exams. The ARC questions require a mix of factual knowledge, commonsense reasoning, and multi-step inference.
    \item The Physical Interaction Question-Answering (PIQA)~\citep{Bisk_Zellers_Lebras_Gao_Choi_2020} benchmark consists of multiple-choice questions about how to accomplish simple physical tasks. Each question presents a short scenario and two possible solutions, between which only one is physically plausible.
    \item The OpenBook QA (OBQA)~\citep{mihaylov-etal-2018-suit} benchmark consists of 6,000 multiple-choice questions based on elementary science facts. Each question is designed to require combining a provided ``open book'' science fact with additional commonsense or general knowledge.
    \item WinoGrande~\citep{Sakaguchi_LeBras_Bhagavatula_Choi_2020} benchmark consists of multiple-choice questions.
Each question presents a sentence with a pronoun and two possible antecedents; the task is to choose the correct referent for the pronoun.
    \item Natural Questions (NQ)~\citep{kwiatkowski-etal-2019-natural} benchmark is designed to evaluate the ability of an AI model to answer real user questions using information from Wikipedia.
    \item The Reading Comprehension from Examinations (RACE)~\citep{Lai2017RACELR} benchmark consists of passages and multiple-choice questions to assess how well AI models can comprehend and reason about written passages.
    \item The Trivia QA benchmark (TQA)~\citep{TQA2017} is a reading comprehension dataset that pairs trivia questions with evidence documents from which answers can be derived.
\end{itemize}

\mypar{Instruction following}
IFEval~\citep{zhou2023ifeval} measures whether a model obeys explicit, programmatically checkable instructions --- constraints such as a required word count, a forbidden word, or a specified output format --- over 541 prompts. Because compliance is verified by a script rather than judged, the metric is objective, but it is only meaningful once a model has been trained to follow instructions, so we report it only for the post-trained models. We report \texttt{inst\_acc}, the fraction of individual constraints satisfied, and \texttt{prompt\_acc}, the fraction of prompts for which every constraint is satisfied.

\mypar{Knowledge and commonsense} The three benchmarks below are reported only for the Qwen3-1.7B experiments, where we probe the knowledge retained by the base model after mid-training. These are not part of the Core-12 average in Table~\ref{tab:benchmarksconfigs}.

\begin{itemize}
    \item The Massive Multitask Language Understanding (MMLU)~\citep{hendrycks2021mmlu} benchmark consists of multiple-choice questions spanning 57 subjects, from elementary mathematics to law and medicine. It is designed to measure the breadth of world knowledge and problem-solving ability acquired during pre-training. We report the macro-average over the 57 subtasks.
    \item The CommonsenseQA (CSQA)~\citep{talmor-etal-2019-commonsenseqa} benchmark consists of multiple-choice questions derived from ConceptNet, where each question has one correct answer and four distractors that share a semantic relation with the same source concept, so that answering requires commonsense knowledge rather than surface cues.
    \item The BoolQ~\citep{clark-etal-2019-boolq} benchmark consists of naturally occurring yes/no questions paired with a Wikipedia passage. Despite the binary answer format, the questions are unprompted and often require entailment-like inference over the passage.
\end{itemize}

\mypar{Models trained from scratch.} In Table~\ref{tab:cast_4T_delta} we report the Core-12 benchmarks for the level 8 models of the scaling ladder, pre-trained for $4\Tchin$ and then mid-trained for a
further $4\Tchin$ followed by $1\Tchin$ of decay. We use a different data mixture during mid-training and we compute the deltas for model casting with respect to the \silu control consisting of exactly the same
mid-training budget without model casting, so that we can measure the cost of model casting alone. That cost is
small: averaged over the twelve benchmarks, \Actcasting loses $0.1$ points at a $75\%$ sparsity
target and $0.4$ at $90\%$, and \LoPcasting loses $0.2$ at $75\%$ and gains $0.1$ at $90\%$. No
individual benchmark moves by more than $2.6$ points in either direction. Note that the two model casting
families disagree on the individual benchmarks that move, this suggests that what is left is largely run-to-run noise rather than a systematic loss. For reference, mid-training itself is worth considerably more than model
casting costs: it lifts the average from $38.2$ to $39.6$ for \Actcasting and from $38.0$ to $39.0$
for \LoPcasting.

\mypar{Why we report extra benchmarks for Qwen3.} Since Qwen3-1.7B-Base is
pre-trained far beyond our scaling-ladder budgets, it retains knowledge that the
ladder models never acquire: level 8 scores $38.2$ on the Core-12 average, close
to chance on the harder multiple-choice tasks, whereas Qwen3-1.7B-Base reaches
$65.0$ on MMLU, $74.9$ on CSQA and $75.6$ on BoolQ. We therefore add these three
knowledge and commonsense benchmarks to the Core-12 suite, to check whether
mid-training erodes what the base model already knows.

\mypar{Qwen3-1.7B.} In Table~\ref{tab:qwen3-delta-5T} we report the downstream performance of an off-the-shelf
checkpoint at the decayed $5\Tchin$ mark (mid-training details in Appendix~\ref{sec:qwen3}), where
the control is the \emph{\silu, no swap} arm: the same mixture and budget, but keeping the native
\silu activation. We again observe that the mid-training mixture contributes to a performance degradation more significantly than model casting. Concretely, it degrades
the Core-12 mean from $51.9$ to $49.4$ and the knowledge-3 mean from $71.8$ to $63.3$, a loss of
$2.5$ and $8.5$ points, whereas applying model casting on top costs only $1.6$ and $0.7$ points
respectively; the activation swap without the $\ell_1$ loss costs $0.4$ on Core-12 while gaining
$1.0$ on knowledge-3. The cost concentrates in code generation, where \Actcasting gives up $4.3$
points on HumanEval+ and $3.8$ on MBPP; on the knowledge benchmarks it loses at most $3.0$ points
and improves on BoolQ. Qualitatively this matches the from-scratch models, e.g. the mixture dominates
and casting is second order.

 \begin{table}[htbp]
     \caption{
     \textbf{Effect of casting mid-trained \texttt{Qwen3-1.7B} downstream performance.}
     Change in downstream performance (percentage points) at the decayed $5T$ checkpoint
      ($100$\,TPP), at peak LR $\gamma=7.40\times10^{-5}$. Deltas are taken against the \emph{\silu, no swap}
      control (the same model mid-trained on the same mixture without the activation swap)
      so they isolate the cost of casting from the cost of the mid-training mixture.
      }
    \centering
    \small

      \centering
      \small
      \setlength{\tabcolsep}{6pt}
      \begin{tabular}{l r rrr}
        \toprule
        & & & \multicolumn{2}{c}{$\Delta$ vs.\ \silu control (points)} \\
        \cmidrule(lr){4-5}
        Task & Qwen3-1.7B-Base & \silu, no swap & R-S+ & \Actcasting \\
             & (pre-trained)   & (control)     & (no $\ell_1$) & @0.90 \\
        \midrule
        \multicolumn{5}{l}{\textit{Core-12 short-context}} \\
        HellaSwag & 66.5 & 69.7 & $-0.7$ & $-1.2$ \\
        ARC-e & 68.1 & 72.7 & $-1.6$ & $-1.6$ \\
        ARC-c & 45.1 & 45.0 & $-0.2$ & $\mathbf{-2.5}$ \\
        PIQA & 76.1 & 76.8 & $+0.3$ & $-0.9$ \\
        WinoGrande & 64.5 & 65.3 & $-0.6$ & $+0.0$ \\
        OBQA & 38.6 & 39.6 & $+0.2$ & $\mathbf{-2.2}$ \\
        RACE-m & 63.9 & 58.4 & $+0.3$ & $-1.0$ \\
        RACE-h & 46.7 & 42.0 & $+0.2$ & $+0.1$ \\
        HumanEval+ & 39.6 & 23.8 & $-1.2$ & $\mathbf{-4.3}$ \\
        MBPP & 55.2 & 40.0 & $-0.6$ & $\mathbf{-3.8}$ \\
        TriviaQA & 38.7 & 40.6 & $-0.7$ & $-1.9$ \\
        NaturalQ & 19.5 & 18.6 & $-0.1$ & $-0.5$ \\
        \cmidrule(lr){1-5}
        \textbf{Core-12 mean} & $\mathbf{51.9}$ & $\mathbf{49.4}$ & $\mathbf{-0.4}$ & $\mathbf{-1.6}$ \\
        \midrule
        \multicolumn{5}{l}{\textit{Knowledge / commonsense}} \\
        MMLU & 65.0 & 57.2 & $-0.8$ & $\mathbf{-3.0}$ \\
        CSQA & 74.9 & 65.6 & $+0.2$ & $\mathbf{-2.0}$ \\
        BoolQ & 75.6 & 67.2 & $\mathbf{+3.5}$ & $\mathbf{+2.8}$ \\
        \cmidrule(lr){1-5}
        \textbf{Knowledge-3 mean} & $\mathbf{71.8}$ & $\mathbf{63.3}$ & $\mathbf{+1.0}$ & $\mathbf{-0.7}$ \\
        \bottomrule
      \end{tabular}
    \label{tab:qwen3-delta-5T}
  \end{table}

  \newcommand{\evalheader}{
  \toprule
  \textbf{Model} & \textbf{HS} & \textbf{WG} & \textbf{ARC-e} & \textbf{ARC-c} & \textbf{PIQA} & \textbf{OBQA} & \textbf{RACE-m} & \textbf{RACE-h} & \textbf{HE+} & \textbf{MBPP} & \textbf{TQA} & \textbf{NQ} & \textbf{Avg} \\
   & acc\_char & acc\_char & acc\_char & acc\_char & acc\_char & acc\_char & acc\_char & acc\_char & pass@1 & pass@1 & f1 & f1 & \\
  \midrule}

\begin{table}[htbp]
  \caption{\textbf{Effect of casting mid-trained level-8 models on downstream performance.} Within each family the deltas are
    taken against that family's \emph{4T} column, that is the same model mid-trained for the same budget
    without casting. The \emph{0T} and \emph{4T} columns are absolute \%: \emph{0T} is the
    pre-trained checkpoint (4T WS + 1T D) before mid-training, so comparing the two gives the gain
    from mid-training alone. All rows are decayed. Bold marks changes of at least 2 points. Per-benchmark metrics are given in Table~\ref{tab:benchmarksconfigs}.
    }
  \label{tab:cast_4T_delta}
  \centering
  \small
  \setlength{\tabcolsep}{4pt}
  \begin{tabular}{l rrrr rrrr}
    \toprule
    & \multicolumn{4}{c}{\Actcasting} & \multicolumn{4}{c}{\LoPcasting} \\
    \cmidrule(lr){2-5} \cmidrule(lr){6-9}
    Task & 0T & 4T & 75\% & 90\% & 0T & 4T & 75\% & 90\% \\
    \midrule
    HS & 57.0 & 59.1 & $-0.4$ & $-1.1$ & 56.7 & 59.1 & $-1.0$ & $-1.6$ \\
    WG & 59.1 & 59.4 & $-0.2$ & $-0.8$ & 57.1 & 58.2 & $-0.3$ & $+1.5$ \\
    ARC-e & 60.3 & 61.5 & $-0.8$ & $-1.4$ & 57.3 & 57.8 & $\mathbf{+2.1}$ & $\mathbf{+2.6}$ \\
    ARC-c & 31.2 & 34.2 & $-1.0$ & $-1.5$ & 31.0 & 31.9 & $+0.6$ & $+0.3$ \\
    PIQA & 72.6 & 73.0 & $+0.5$ & $+0.5$ & 73.0 & 74.0 & $-0.5$ & $-0.5$ \\
    OBQA & 34.2 & 35.8 & $+0.2$ & $+1.0$ & 36.6 & 36.0 & $+0.6$ & $-0.8$ \\
    RACE-m & 52.2 & 53.5 & $-0.2$ & $-0.8$ & 50.6 & 52.3 & $-0.3$ & $+0.1$ \\
    RACE-h & 38.3 & 39.5 & $-0.5$ & $-0.6$ & 38.4 & 39.2 & $+0.4$ & $-0.2$ \\
    HE+ & 6.7 & 6.7 & $\mathbf{+2.4}$ & $+1.8$ & 7.9 & 7.3 & $-1.2$ & $+1.8$ \\
    MBPP & 10.8 & 13.0 & $-0.2$ & $-0.4$ & 11.6 & 12.2 & $-1.4$ & $+0.4$ \\
    TQA & 24.1 & 26.4 & $+0.0$ & $-0.6$ & 24.3 & 27.0 & $-0.6$ & $-1.3$ \\
    NQ & 11.6 & 12.6 & $-0.7$ & $-0.6$ & 11.5 & 13.1 & $-0.7$ & $-1.3$ \\
    \cmidrule(lr){1-9}
    \textbf{Average} & \textbf{38.2} & \textbf{39.6} & $-0.1$ & $-0.4$ & \textbf{38.0} & \textbf{39.0} & $-0.2$ & $+0.1$ \\
    \bottomrule
  \end{tabular}
\end{table}

\section{Post-training}
\label{sec:post-training}
\mypar{LCFT and SFT.} To check that model casting survives post-training, we use our own mid-trained level 8 models (starting from the corresponding pre-trained models from the scaling ladder of Table~\ref{tab:ladder}), with and without model casting, and apply a standard post-training
pipeline: long-context fine-tuning (LCFT) followed by supervised fine-tuning
(SFT). Six variants are compared
throughout: \Actcasting and \LoPcasting at
$75\%$ and $90\%$ sparsity targets versus a dense control with \silu.

\begin{table}[htbp]
\centering
\caption{Average downstream performance across post-training, relative to the \silu dense baseline. $\Delta_{\mathrm{pre}}$ is the gap before
post-training, $\Delta_{\mathrm{SFT}}$ the gap after $8000$ SFT steps, and
$\Delta_{\mathrm{corr}}\,{=}\,\Delta_{\mathrm{SFT}}-\Delta_{\mathrm{pre}}$
isolates what post-training itself costs each variant. All values are rounded to one
decimal, so the columns reconcile exactly.}
\label{tab:postrain-avg}
{\small
\begin{tabular}{@{}lrrr@{\quad}rrr@{}}
  \toprule
  & \multicolumn{3}{c}{Core-12 average}
  & \multicolumn{3}{c}{relative to the dense baseline} \\
  \cmidrule(lr){2-4} \cmidrule(l){5-7}
  Setup (pre-training $\to$ mid-training)
    & pre-LCFT & LCFT & SFT
    & $\Delta_{\mathrm{pre}}$
    & $\Delta_{\mathrm{SFT}}$
    & $\Delta_{\mathrm{corr}}$ \\
  \midrule
  \Base[\silu{}] $\to$ \Base[\silu{}]                   & 39.6 & 39.7 & 36.6 & --- & --- & --- \\
  \addlinespace
  \Base[\silu{}] $\to$ \Base[\RminSplus, $\ell_1$@75\%] & 39.5 & 39.0 & 36.8 & \dneg{$-0.1$} & \dpos{$+0.2$} & \dpos{$+0.3$} \\
  \Base[\silu{}] $\to$ \Base[\RminSplus, $\ell_1$@90\%] & 39.2 & 38.7 & 36.0 & \dneg{$-0.4$} & \dneg{$-0.6$} & \dneg{$-0.2$} \\
  \addlinespace
  \LoPA[\silu{}] $\to$ \LoPA[\RminSplus, $\ell_1$@75\%] & 38.8 & 38.6 & 36.1 & \dneg{$-0.8$} & \dneg{$-0.5$} & \dpos{$+0.3$} \\
  \LoPA[\silu{}] $\to$ \LoPA[\RminSplus, $\ell_1$@90\%] & 39.1 & 38.1 & 36.0 & \dneg{$-0.5$} & \dneg{$-0.6$} & \dneg{$-0.1$} \\
  \bottomrule
\end{tabular}}
\end{table}

Table~\ref{tab:postrain-avg} gives the Core-12 evaluation tasks average relative to the dense
baseline, before post-training and after $8000$ SFT steps. After SFT, the model casting variants that target 90\% sparsity have a small degradation of 0.1--0.2 with respect to the dense baseline. However, both \Actcasting and \LoPcasting @75\% improve by 0.3 after SFT with respect to the dense baseline. In contrast, the LCFT stage is close to
neutral, moving the average by at most $1.0$ point in either direction, so SFT
accounts for almost all of the change. SFT is expensive for every model alike,
costing $2.4$ to $3.2$ points by step $8000$. Hence, the degradation is due to the SFT datamix rather than model casting.

We also run a suite of chat-format evaluations, of which three carry
positive instruction-following signal at this scale: \texttt{cruxeval\_output}, \texttt{gsm8k\_chat} and
\texttt{mbpp\_chat}. Across the six variants, these agree to within $1.6$, $4.0$
and $3.2$ points at step $8000$, so model casting does not measurably change
chat behavior. In particular, on \texttt{gsm8k\_chat} the chat model beats the base model
by a wide margin ($7.8$ against $4.9$), confirming that SFT trains the chat
template. We do not report the multiple-choice chat evaluations, which sit at or
below chance, nor a format-compliance check that saturates on every variant.
Finally, in Table~\ref{tab:cast_postrain_ifeval}, we report instruction-following abilities on IFEval~\citep{zhou2023ifeval}
after $8000$ SFT steps. Note that we can only conduct this evaluation after SFT has
trained the chat template. Both pre-training and LCFT use the plain
\texttt{tiktoken} tokenizer. Every $\Delta$ is measured against the single
dense control in the first row, so the \LoPcasting entries also carry the
architecture difference. We can conclude that all model casting variants retain measurable instruction-following ability. In particular, the \LoPcasting at 75\% does better than the dense baseline. There is some degradation for the 90\% model casting variants which could be attributed to using such an extreme sparsity operating point.

\begin{table}[htbp]
\centering
\caption{IFEval after $8000$ SFT steps. Scores are \emph{loose}; the
strict variant penalizes formatting alone and is largely common-mode across
variants. $\Delta$ is measured against the \Actcasting dense control in the first
row, a single reference for the whole table, so the \LoPcasting rows also carry
the architecture difference.}
\label{tab:cast_postrain_ifeval}
{\small\setlength{\tabcolsep}{5pt}
\begin{tabular}{@{}lrrrr@{}}
\toprule
& \multicolumn{2}{c}{\texttt{inst\_acc}} & \multicolumn{2}{c}{\texttt{prompt\_acc}} \\
\cmidrule(lr){2-3} \cmidrule(lr){4-5}
Setup (pre-training $\to$ mid-training) & score & $\Delta$ & score & $\Delta$ \\
\midrule
\Base[\silu{}] $\to$ \Base[\silu{}]                   & 33.9 & ---              & 22.9 & ---              \\
\Base[\silu{}] $\to$ \Base[\RminSplus, $\ell_1$@75\%] & 33.9 & $+0.0\%$         & 22.0 & \dneg{$-3.9\%$}  \\
\Base[\silu{}] $\to$ \Base[\RminSplus, $\ell_1$@90\%] & 32.0 & \dneg{$-5.6\%$}  & 21.1 & \dneg{$-7.9\%$}  \\
\addlinespace
\LoPA[\silu{}] $\to$ \LoPA[\RminSplus, $\ell_1$@75\%] & 35.1 & \dpos{$+3.5\%$}  & 23.3 & \dpos{$+1.7\%$}  \\
\LoPA[\silu{}] $\to$ \LoPA[\RminSplus, $\ell_1$@90\%] & 32.5 & \dneg{$-4.1\%$}  & 19.2 & \dneg{$-16.2\%$} \\
\bottomrule
\end{tabular}}
\end{table}

\end{document}